\documentclass[11pt, letterpaper]{article}
\usepackage{acmlab}
\usepackage{url}              

\usepackage{float}
\usepackage[section]{placeins}  
\usepackage{array}            
\usepackage{pifont}           

\begin{document}

\acmlabtitle{NeuroPilot: An Agent-Driven Smart Pipeline for Processing, Quality Control, and Managing Neuroimages}
  {Yiyao Chen\equalmark, Yucheng Li\equalmark, Junhong Tong, Shaoqi Wang, Kunhao Zhou, Ziquan Wei, \\
   Monica Murea, Ran Ren, Marissa DiPiero, Tingting Dan\corrmark, Guorong Wu\corrmark}
  {Departments of Psychiatry and Computer Science \\
   University of North Carolina at Chapel Hill}
  {}
\vspace{-1.5em}
\centerline{\small \corrmark Corresponding authors: \href{mailto:tingting_dan@med.unc.edu}{\{tingting\_dan,grwu\}@med.unc.edu}}
\equalstatement
\vspace{1em}

\begin{abstract}
\noindent Transforming raw neuroimage {(such as T1w, fMRI and DWI)} archives into analysis-ready derivatives relies on three brittle stages: \textit{data standardization}, \textit{modality-specific preprocessing}, and \textit{quality control (QC)}. While individual neuroimaging tools are well developed, their orchestration requires project-specific scripts, environment-adaptive tuning, and labor-intensive manual QC. To address this, we introduce \textbf{NeuroPilot}, a multi-agent system that digitalizes the expertise of neuroimage processing, quality control, and data management into three LLM-invocable skills: \texttt{dcm2bids-skill}, \texttt{neuroimage-pre-skill}, and \texttt{qc-agent-skill}. The LLM-driven agent autonomously orchestrates workflows, generalizing various infrastructure settings into a single configuration to achieve the highest scalability. Demonstrating the system's generalizability, we deployed \textbf{NeuroPilot} across 17 diverse cohorts (\textbf{>123,000} subjects) spanning infant to aging populations and multiple MRI modalities (structural, diffusion, functional, perfusion). In practice, after standardizing data via the \texttt{dcm2bids-skill}, the agent dynamically routes datasets to the optimal \texttt{neuroimage-pre-skill} based on available modalities and cohort traits (e.g., dispatching T1w and fMRI data to fMRIPrep, or selecting specialized pipelines for infant cohorts). The \texttt{qc-agent-skill} then drives an evidence-based, semi-automated QC via a 3-D browser dashboard, utilizing a multi-tiered verification system to optimize failed cases and escalate complex issues for supervisor inspection. Quantitatively, our QC agent screened 4,122 production subjects, validating its automated flags against FreeSurfer's topology-defect metrics. The infant processing pipeline achieved a 97.3\% completion rate on QC-validated inputs. Importantly, \textbf{NeuroPilot} compresses the traditional 2--3 month timeline for training staff and processing complete datasets into a single week. \textbf{NeuroPilot} is deployed at \url{https://wanda-cyberbench.com/}.
\end{abstract}

\section{Introduction}

Reproducible brain-MRI research relies on reliable computational tools for neuroimages. For example, data standardization uses \texttt{dcm2niix}/\texttt{dcm2bids}~\citep{dcm2niix} over the Brain Imaging Data Structure (BIDS)~\citep{bids}. Preprocessing uses fMRIPrep~\citep{fmriprep}, sMRIPrep, QSIPrep/QSIRecon~\citep{qsiprep}, XCP-D~\citep{xcpd}, FreeSurfer~\citep{freesurfer}, and ANTs~\citep{ants}, packaged as containers~\citep{singularity}. Meanwhile, quality control (QC) draws on MRIQC image-quality metrics~\citep{mriqc} and practices such as surface topology-defect screening~\citep{euler}. Each tool, on its own, is well documented and validated in many neuroimaging studies.

Taking a raw archive all the way to quality-controlled imaging traits, however, is not one-stop shopping. Successful batch processing neuroimages consists of multiple steps, each relying on inter-dependent data processing and QC. For example, data standardization faces compressed folders of extension-less DICOMs, vendor-specific \texttt{SeriesDescription} strings, and protected health information (PHI). Manual effort is needed to locate the folder of DICOMs, map each series to the right BIDS datatype and suffix, handle multiple sessions, catch modalities present in only a few subjects, and validate the output. After that, image preprocessing relies on experts to read off available modalities, pick the pipeline, feed intermediate output from FreeSurfer into the next step, write process scripts with the correct resources and bind mounts, batch across hundreds of subjects, reclaim scratch space, and confirm that every stage produced what it should. Quality control demands human labor to inspect each intermediate, decide which subjects fail, apply or reject fixes, and document the processing log.

Multiple domain-specific workflow and complexity of real-world data are the major reasons why fully automation fails in practice, which hinder the efficiency and replicability of neuroimaging studies. Image processing script is often specifically designed for each project, tuned to a specific computational environment, lack of sufficient QC, and rarely structured for software reuse. Quality control faces a more critical challenge: it is often labor-intensive, subjective, and undocumented. In this regard, the orchestration of existing tool is the real crux. 

\begin{figure}[h]
\centering
\includegraphics[trim=0cm 0cm 0cm 0cm, clip, width=1\linewidth]{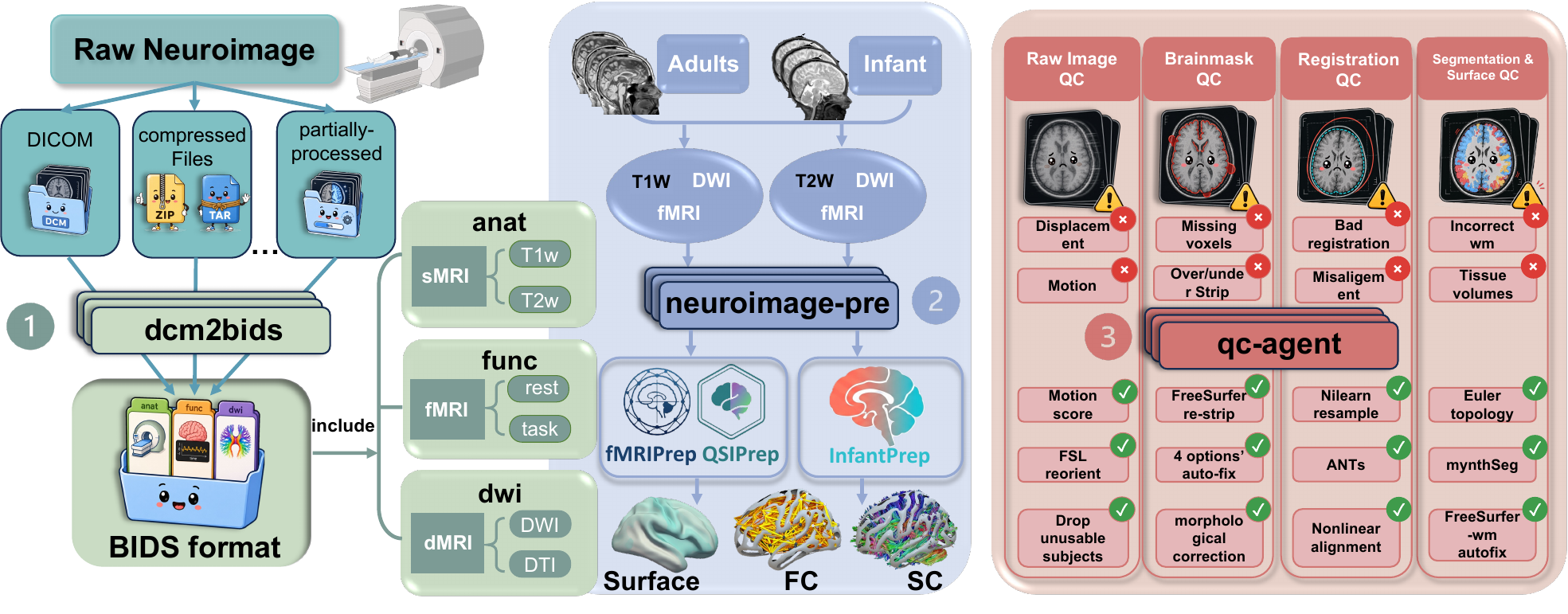}
\vspace{-1.5em}
\caption{\textbf{Overview of our \textbf{NeuroPilot}.} \ding{182}\texttt{dcm2bids-skill} standardizes raw DICOM into BIDS, \ding{183}\texttt{neuroimage-pre-skill} preprocesses it to FC/SC/surface derivatives, and \ding{184}\texttt{qc-agent-skill} reviews the image processing results. Automated checkpoints sit between stages, and human-approval gates guard every decision.}
\label{fig:overview}
\end{figure}

It is clear that the complete workflow of high-quality image processing is not only massively time-consuming, but also in a high demand of teamwork \footnote{It has been frequently reported that team personnel change leads to project delay, lost of traceability, and rendering anomalous results impossible to track down.}. 
To address this challenge, we use agentic AI technology \citep{ren2026cyberneuro} to read a filesystem, run shell commands, call tools, and apply reasoning over the results by alternating reasoning with action~\citep{react} and learning when to invoke a tool~\citep{toolformer}. 
By packaging the operational knowledge (such as tool selection, execution sequence, input parameters, and validation criteria) into an agent-executable format, we offload repetitive routine work to the agent while allowing researchers to put more focus on scientific thinking. In this context, we generalize years' of successful neuroimage processing expertise into a multi-agent system called \textbf{NeuroPilot}, with three declarative agent skills (as shown in Fig. \ref{fig:overview}). Each skill is a self-describing capability package with four components: (1) a natural-language description that tells the agent when it applies, (2) reference procedures read on demand, (3) parameterized scripts that take all paths as arguments, and (4) checkpoints that verify inputs before a run and outputs after. Serving as the foundation of a comprehensive neuroimaging pipeline, these skills encompass the full processing lifecycle, tailored for each stage: a data-standardization skill (\texttt{dcm2bids-skill}), a modality-specific preprocessing skill (\texttt{neuroimage-pre-skill}, and a human-in-the-loop QC skill (\texttt{qc-agent-skill}). The agent chains them from context and the human supervises the consequential steps. Our contributions include:

\begin{itemize}
\item \textbf{An end-to-end agentic design} for the whole neuroimage processing lifecycle (data standardization, modality-specific preprocessing, and QC) as three declarative, LLM-invocable skills withinput/output checkpoints, separating portable domain logic from site-specific configurations for universal cross-platform execution.
\item \textbf{A reproducible, human-in-the-loop QC stage} that dynamically grades subjects against their own cohort using published metrics (e.g., Iglewicz--Hoaglin modified~$Z$ and SynthStrip agreement). Furthermore, it visualizes results via integrated 3-D browser dashboards (e.g., \texttt{niivue} for .nii format data), and decouples error detection from repair mechanisms, and records every interaction in an append-only audit ledger.
\item \textbf{Large-scale deployment and validation} testing over 123,000 subjects from 17 cohorts. This platform verified the agent's capability for automated series classification, dynamic pipeline selection, and idempotent batch processing across $\sim$20 atlases in high-throughput settings, incorporating a five-step structural QC with pre-calibrated grading thresholds.
\item \textbf{A substantial reduction in time cost and human labor.} By delegating repetitive pipeline orchestration to the agent and filtering out trivial checkpoints, this framework compresses the end-to-end data processing timeline from months to a single week, accelerating large-scale neuroimaging research.
\end{itemize}

\section{Method}
\subsection{Materials}
\label{supp:atlase}
\textbf{Multi-cohort validation setup.}
To date, we selected samples from seventeen publicly or institutionally available cohorts spanning infant, developmental, young-adult, and aging/neurodegeneration populations (involving more than 123{,}000 subjects in total, Fig. \ref{fig:result}a), chosen to vary in vendor, field strength, protocol, and available modalities so that series classification and pipeline selection are not trivial. The \texttt{dcm2bids-skill}  converted every dataset from raw DICOM to BIDS, and the BIDS Validator checked each one before image preprocessing step.

\begin{table}[!h]
\centering
\caption{Validation datasets. \emph{Total $N$}: number of participants available per cohort. Filled circles ($\bullet$) mark modalities the cohort offers under its standard protocol (Note: T1w/T2w/FLAIR grouped under anat folder).}
\label{supp:tab:testbed}
\footnotesize
\begin{tabular}{l r c c c c l}
\toprule
\textbf{Cohort} & \textbf{Total $N$} & \textbf{anat} & \textbf{dwi} & \textbf{func} & \textbf{perf} & \textbf{Notes} \\
\midrule
ADNI        & 4{,}122   & $\bullet$ & $\bullet$ & $\bullet$ & $\bullet$ & CN/MCI/AD; longitudinal \\
ADNI-DOD    & 414       & $\bullet$ & $\bullet$ & $\bullet$ &           & Vietnam-veteran cohort \\
AIBL        & 1{,}538   & $\bullet$ & $\bullet$ &           &           & Aging + AD; PET \\
BLSA (open) & 118       & $\bullet$ & $\bullet$ &           &           & Lifespan aging \\
HABS-HD     & 6{,}540   & $\bullet$ & $\bullet$ & $\bullet$ &           & Health-disparities cohort \\
MCSA        & 2{,}122   & $\bullet$ & $\bullet$ &           &           & Population aging \\
NIFD        & 346       & $\bullet$ & $\bullet$ &           &           & FTD (bvFTD/PPA) \\
POINTER     & 1{,}008   & $\bullet$ & $\bullet$ &           & $\bullet$ & Lifestyle-intervention imaging \\
PPMI        & 10{,}277  & $\bullet$ & $\bullet$ & $\bullet$ &           & PD/HC/prodromal \\
SCAN        & 9{,}295   & $\bullet$ & $\bullet$ & $\bullet$ & $\bullet$ & NACC multimodal aggregation \\
UK Biobank  & 71{,}852  & $\bullet$ & $\bullet$ & $\bullet$ &           & Population imaging \\
WRAP        & 816       & $\bullet$ & $\bullet$ &           &           & Alzheimer's-prevention registry \\
HCP-YA      & 1{,}200   & $\bullet$ & $\bullet$ & $\bullet$ &           & Young-adult connectome \\
HCP-A       & 717       & $\bullet$ & $\bullet$ & $\bullet$ &           & Aging connectome \\
BIOCARD     & 744       & $\bullet$ &           & $\bullet$ &           & Preclinical AD \\
ABCD        & 11{,}617  & $\bullet$ & $\bullet$ & $\bullet$ &           & Adolescent development \\
EBDS        & 444       & $\bullet$ & $\bullet$ & $\bullet$ &           & Infant; CONTE2 (213) + TWINS2 (231) \\
\bottomrule
\end{tabular}
\end{table}

\textbf{QC threshold calibration.}
The QC grading thresholds are strictly cohort-relative: each subject is evaluated against the empirical distribution of its own cohort. For example, a subject is flagged if its normalized cross-correlation (NCC) registration falls below the mean of the cohort minus $2\sigma$ (where $\sigma$ represents the standard deviation), or if its surface topology-defect count exceeds the median of the cohort plus $k \cdot \text{MAD}$ (where $\text{MAD}$ denotes the absolute median deviation and $k$ is a predefined scaling factor). These relative metrics are backed by conservative absolute floors as a safety net. Both the relative cutoffs and the absolute floors were fixed globally and applied identically across all cohorts, ensuring that no per-cohort site tuning biases the reported results.

\textbf{Atlases and derivatives.}
The pipeline computes functional connectivity (FC), structural connectivity (SC), and cortical thickness derivatives across 19 standard adult and neonatal parcellations (e.g., Schaefer\citep{schaefer}, AAL~\citep{aal}) and multiple spatial resolutions. Functional connectivity (FC) matrices are computed in
\texttt{MNI152NLin2009cAsym} space across 19 parcellations. Fourteen come from
XCP-D~\citep{xcpd}: the ten \texttt{4S} hybrid atlases
(\texttt{4S156Parcels}--\texttt{4S1056Parcels}, each pairing a Schaefer cortical
solution of 100--1000 parcels~\citep{schaefer} with a fixed set of 56
subcortical, thalamic and cerebellar parcels), Glasser/HCP-MMP
(360;~\citep{glasser}), Gordon (333;~\citep{gordon}), HCP subcortical, and
Tian~\citep{tian}. Five more are applied post hoc with
\texttt{nilearn}~\citep{nilearn}: AAL (116;~\citep{aal}), Destrieux
(160;~\citep{destrieux}), Brainnetome (246;~\citep{brainnetome}), AICHA
(384;~\citep{aicha}) and Shen (268;~\citep{shen}). Each run yields a
region$\times$region correlation matrix together with the underlying region-wise
time series, both in CSV. Structural connectivity (SC) matrices are written by
QSIRecon as MATLAB \texttt{.mat} files for five parcellations (AAL116,
AICHA384Ext, Brainnetome246Ext, Gordon333Ext and a 161-region Destrieux-derived
atlas), while neonatal data use two, the AAL and dHCP parcellations. Cortical
thickness is retained at native FreeSurfer resolution~\citep{freesurfer} and
resampled to \texttt{ico7} (163{,}842 total vertices, i.e., 81{,}921 per
hemisphere), \texttt{ico5} (10{,}242 vertices per hemisphere), and \texttt{ico4}
(2{,}562 vertices per hemisphere).

\textbf{Skills as the unit of capability.} A skill comprises four components: a natural-language \textbf{description} for context-aware agent selection, on-demand \textbf{reference procedures} to minimize context size, \textbf{parameterized scripts} accepting dynamic paths, and \textbf{checkpoints} for input/output verification. Our framework implements three sequential skills (\texttt{dcm2bids-skill}, \texttt{neuroimage-pre-skill}, \texttt{qc-agent-skill}), where data standardization feeds preprocessing, which in turn feeds quality control. Underlying neuroimaging tools (e.g., fMRIPrep, FreeSurfer, MRtrix3) are abstracted as implementation details managed autonomously by the agent.

\textbf{Portable logic vs.\ environment configuration.} A single \texttt{pipeline.env} file, generated from a template, centralizes all site-specific settings: container image paths, licenses, atlas directories, \texttt{APPTAINER\_BIND} mounts, module names, and execution backends (e.g., local, server, or SLURM). Because the skills reference only these variables, migrating the entire pipeline to a new computational environment requires editing just this one file.

\textbf{Checkpoints.} A unified \texttt{check\_status.sh} script enforces dual validations at each stage. The pre-run check verifies that prerequisite inputs exist, aborting early to prevent wasteful computation. The post-run check scans filesystem outputs or reads success sentinels from job logs to report per-subject outcomes. By empirically verifying the presence of expected artifacts, these checkpoints ground the agent's decision-making in actual system state rather than assumed success. Additionally, the data standardization stage incorporates a \emph{drop-check} mechanism to flag any imaging modalities silently omitted across a cohort.

\textbf{The agent control loop.}
Given a dataset and a goal, the agent inspects the filesystem, standardizes to BIDS, reads off the modalities and preprocesses, drives QC, reports per-subject outcomes, and resubmits failures idempotently. It escalates to the human at consequential steps. Deleting or overwriting data, launching large job arrays, choosing a fallback tool or a modeling recipe, dismissing a subject, and exporting QC results all need explicit approval. The QC checkpoints exist so the human sees verified state, that is, which subjects actually succeeded, rather than the agent's word for it.

\subsubsection{Data standardization (\texttt{dcm2bids-skill})}
\label{sec:ingest}
\begin{figure}[!h]
\centering
\includegraphics[width=\textwidth]{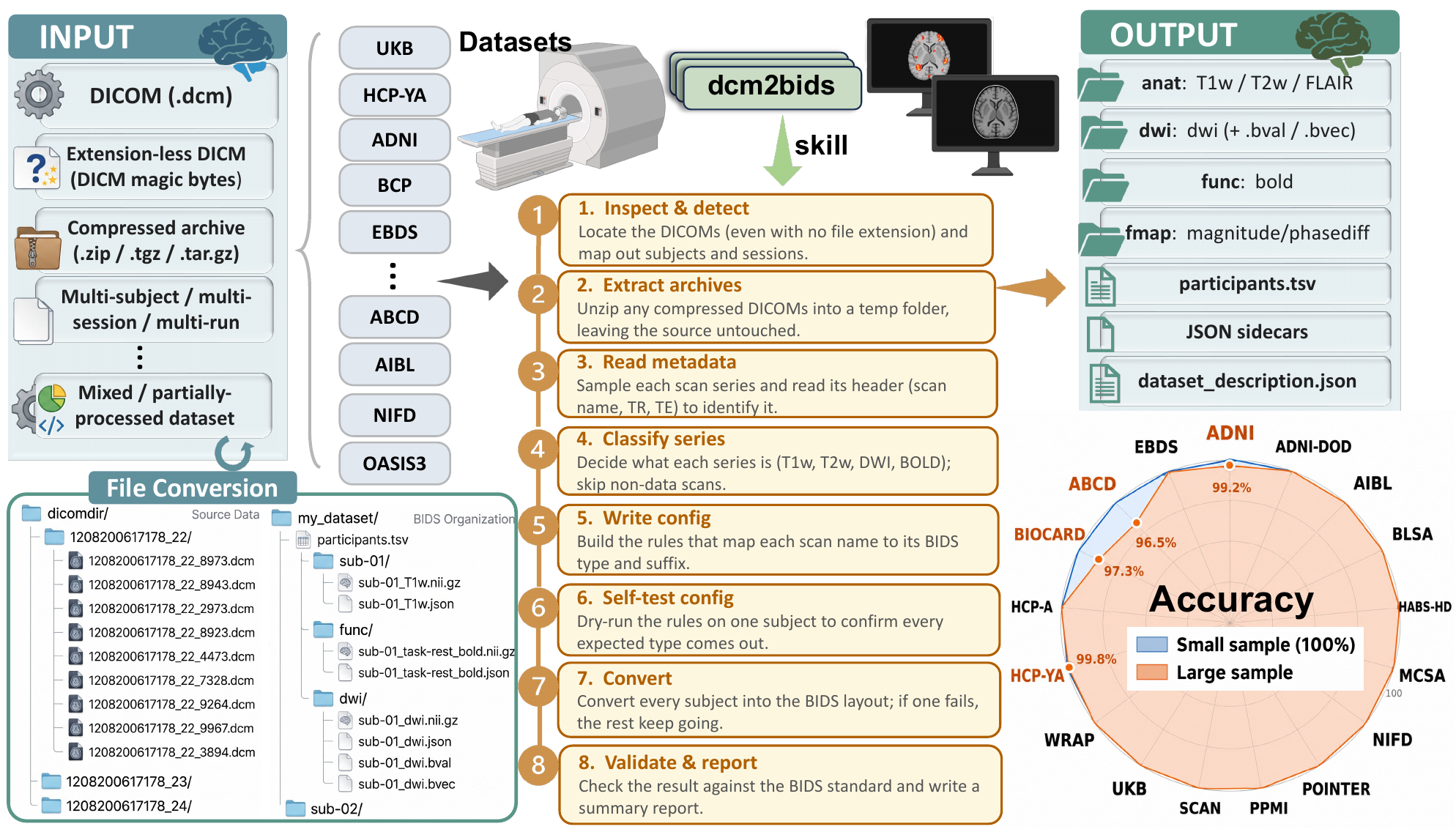}
\caption{\textbf{Data-standardization stage -- \texttt{dcm2bids-skill}.} Heterogeneous inputs (extension-less DICOM found by magic bytes, \texttt{.zip}/\texttt{.tgz} archives, mixed or partially processed datasets) become a validated BIDS dataset. The engine stays \texttt{dcm2niix}/\texttt{dcm2bids}; the skill adds inspection, classification, self-testing, and validation around it.}
\label{fig:dcmbids2}
\end{figure}

The \texttt{dcm2bids-skill} converts raw DICOM acquisitions into a BIDS dataset. The agent derives the \texttt{dcm2bids} configuration from the data, so users no longer have to write this error-prone file by hand. To cope with messy real-world exports, the agent identifies DICOM files by the \texttt{DICM} signature at byte offset 128, not by file extension, and unpacks compressed archives into temporary workspaces. A configuration built from a single subject can miss modalities that only some subjects have, such as DWI or ASL. The agent therefore takes the union of distinct series across the whole cohort and builds one batch configuration from it. To map each series to a BIDS suffix, it runs \texttt{dcm2bids\_helper} on a representative subject and reads sidecar fields such as \texttt{SeriesDescription} and \texttt{Modality}. Before converting anything, the agent bundles all pending decisions, including label normalization, session labels, and execution mode (e.g., \texttt{sbatch}, \texttt{nohup}), into one request for the user to confirm, so the run does not stall on repeated prompts. The same request lists any protected health information (PHI) the agent has found. Conversion starts only after the user confirms de-identification. After conversion, the skill runs the BIDS Validator, writes \texttt{conversion\_report.md}, and performs a \emph{drop-check} to confirm that no expected modality was lost. The skill converted all seventeen testbed cohorts, and every resulting dataset passed BIDS validation (Fig.~\ref{fig:result}(a,b) and Table~\ref{supp:tab:testbed}, workflow in Fig.~\ref{fig:dcmbids2}).

\begin{figure}[!h]
\centering
\includegraphics[trim=0cm 0cm 0cm 0cm, clip, width=\linewidth]{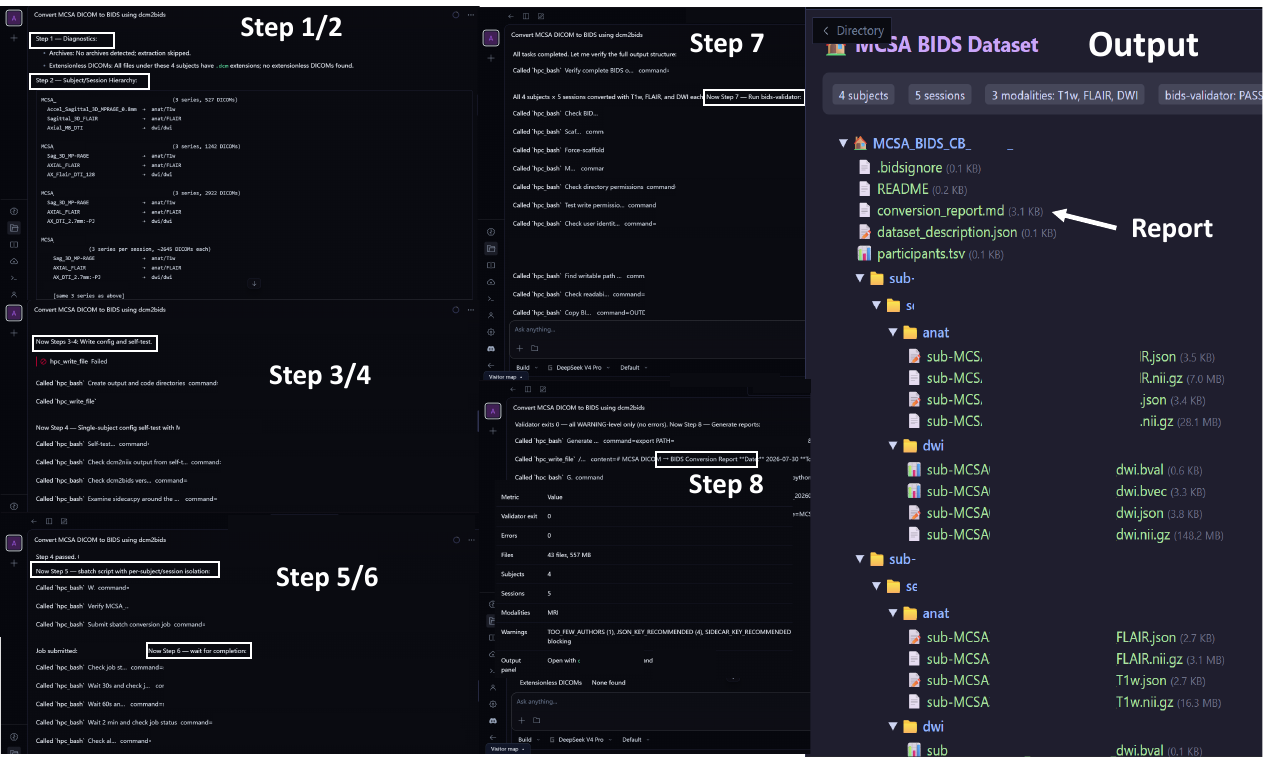}
\vspace{-1.5em}
\caption{\textbf{How \texttt{dcm2bids-skill} works.} }
\label{fig:supp-dcm2bids-steps}
\end{figure}

\textbf{Step-by-step workflow.} As illustrated in Fig.~\ref{fig:dcmbids2} and Fig.~\ref{fig:supp-dcm2bids-steps}, the data standardization stage processes heterogeneous inputs into a validated BIDS dataset through the following automated steps:

\begin{itemize}
    \item \textbf{Step 1: Inspect \& detect.} The agent locates the DICOM files (including extension-less DICOMs identified by magic bytes) and maps out the subject and session hierarchies.
    \item \textbf{Step 2: Extract archives.} Any compressed archives (\texttt{.zip}, \texttt{.tgz}) are extracted into a temporary directory, ensuring the source data remains untouched.
    \item \textbf{Step 3: Read metadata.} The agent samples each scan series and reads its DICOM header (e.g., scan name, TR, TE) for identification.
    \item \textbf{Step 4: Classify series.} Based on the metadata, the agent categorizes each series (e.g., T1w, T2w, DWI, BOLD) and filters out non-data scans (e.g., localizers).
    \item \textbf{Step 5: Write config.} The system dynamically builds the JSON configuration rules mapping each scan name to its appropriate BIDS datatype and suffix.
    \item \textbf{Step 6: Self-test config.} A dry-run is executed on a single subject to confirm that the configuration produces the expected BIDS outputs without errors.
    \item \textbf{Step 7: Convert.} The \texttt{dcm2bids} engine is executed across all subjects. The pipeline is robust; if one subject fails, the conversion continues for the others.
    \item \textbf{Step 8: Validate \& report.} The final output is checked against the BIDS standard validator, and a comprehensive summary report is generated.
\end{itemize}

\noindent
\subsubsection{ Modality-specific preprocessing
(\texttt{neuroimage\allowbreak-pre\allowbreak-skill})}
\begin{figure}[!h]
\centering
\includegraphics[trim=0cm 0cm 0cm 0cm, clip, width=\linewidth]{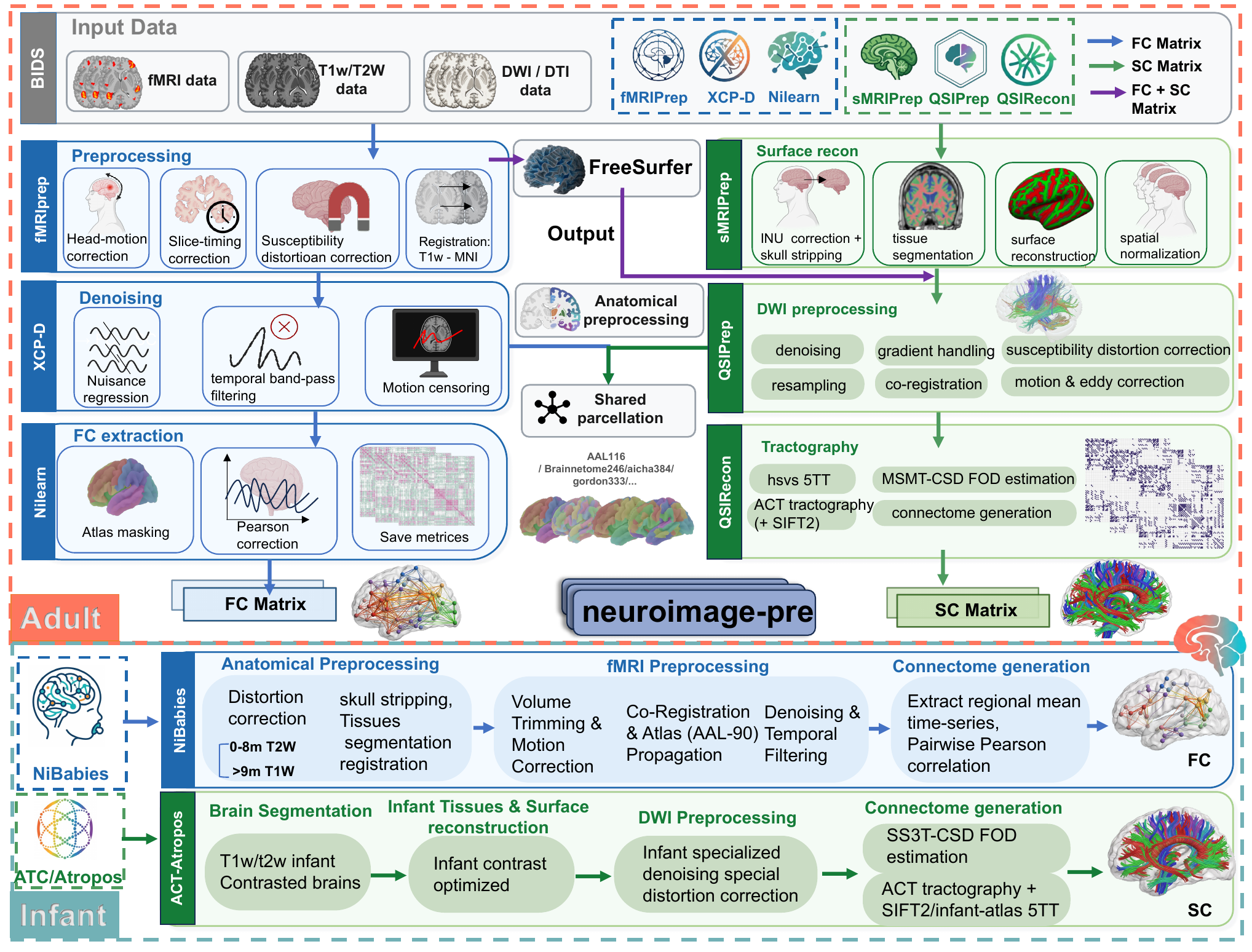}
\caption{\textbf{Modality-specific preprocessing  -- \texttt{neuroimage-pre-skill}.} The skill routes input data (fMRI, T1w/T2w, DWI) through dedicated adult or infant pipelines. For adult cohorts, the functional arm chains fMRIPrep, XCP-D, and Nilearn to extract functional connectivity (FC) matrices, while the structural arm runs sMRIPrep, QSIPrep, and QSIRecon (featuring 5-tissue-type segmentation, MSMT-CSD, and ACT with SIFT2) for structural connectivity (SC). For infant cohorts, the pipeline leverages NiBabies for neonatal functional processing and ACT-Atropos alongside SS3T-CSD for infant-optimized structural tractography. Both populations share unified FreeSurfer surface reconstructions and common parcellation schemes.}
\label{supp:fig:preproc}
\end{figure}

This skill picks one of five pipelines by detected modality (Table~\ref{tab:select}) and drives the matching BIDS Apps, loading only the reference it needs. The full workflow of this skill is shown in Fig.~\ref{supp:fig:preproc}.

\begin{table}[h]
\centering
\small
\setlength{\tabcolsep}{2pt}
\renewcommand{\arraystretch}{1.15}
\caption{\textbf{Modality-driven pipeline selection.}
The agent dynamically parses available input modalities and cohort
demographics (adult (top) vs.\ infant (bottom)) to route data through the appropriate
functional or structural pipelines, ultimately generating standardized
FC and SC derivatives.}\label{tab:select}
\begin{tabular}{
    @{}
    >{\raggedright\arraybackslash}p{0.25\columnwidth}
    >{\raggedright\arraybackslash}p{0.54\columnwidth}
    >{\centering\arraybackslash}p{0.11\columnwidth}
    @{}
}
\toprule
\textbf{Cohort \& Modality}
& \textbf{Pipeline Workflow}
& \textbf{Output} \\
\midrule
fMRI + T1w
& fMRIPrep $\to$ XCP-D $\to$ FC extraction
& FC \\

DWI + T1w
& sMRIPrep $\to$ QSIPrep $\to$ QSIRecon
& SC \\

fMRI + DWI + T1w
& fMRIPrep $\to$ XCP-D $\to$ FC extraction, 
QSIPrep $\to$ QSIRecon
& FC + SC \\

\midrule
fMRI + T1w/T2w
& NiBabies $\to$ custom FSL/ANTs pipeline
$\to$ FC extraction
& FC \\

DWI + T1w/T2w
& QSIPrep $\to$ ACT-Atropos tractography
$\to$ {tck2connectome}
& SC \\
\bottomrule
\end{tabular}
\end{table}


\textbf{Structural MRI + fMRI + DWI $\to$ FC + SC.} For adult cohorts with multimodal data, the pipeline orchestrates parallel functional and structural workflows. fMRIPrep~\citep{fmriprep} performs the foundational anatomical and BOLD preprocessing, including FreeSurfer surface reconstruction. The functional stream then passes to XCP-D~\citep{xcpd} for rigorous denoising (nuisance regression, band-pass filtering, and motion censoring) before nilearn~\citep{nilearn} computes the FC matrices. Concurrently, the structural arm leverages the FreeSurfer surfaces to initialize QSIPrep~\citep{qsiprep} diffusion preprocessing. Finally, QSIRecon utilizes MRtrix3~\citep{mrtrix} to execute MSMT-CSD~\citep{msmt}, anatomically-constrained tractography~\citep{act} guided by HSVS 5-tissue-type segmentation, and SIFT2 streamline weighting~\citep{sift2}, ultimately extracting SC matrices across $\sim$20 atlases (Section~\ref{supp:atlase}).

\textbf{Adult structural MRI + DWI $\to$ SC.} If we only have structural MRI + DWI data, the agent routes structural inputs through sMRIPrep to generate the prerequisite FreeSurfer reconstructions. QSIPrep and QSIRecon then leverage these surfaces to execute the automated \texttt{mrtrix\_multishell\_msmt\_ACT-hsvs} workflow on the DWI data (incorporating a single-shell override where necessary). Note, if an sMRIPrep container is locally unavailable, the agent autonomously falls back to fMRIPrep in \texttt{--anat-only} mode and standardizes the output paths.

\textbf{Infant FC and SC variants.}
Neonatal tissue contrast fundamentally breaks standard adult templates, requiring dedicated age-adaptive workflows. For functional connectivity, the infant route leverages NiBabies for specialized anatomical and BOLD preprocessing, followed by a custom FSL/ANTs pipeline for rigorous denoising and FC matrix extraction. For structural connectivity, the pipeline utilizes QSIPrep for diffusion preprocessing and an ACT-Atropos workflow for tractography. This SC route registers age-matched infant atlases (UNC neonatal/1-year/2-year, BCP monthly, dHCP, ENA50 and M-CRIB) into subject T2 space with ANTs SyN, builds an infant-optimized 5-tissue-type image, computes SS3T FODs, and runs 10M-streamline ACT tractography with SIFT2 to write SC matrices over ten parcellations spanning those five atlas families, from UNC AAL-90 to JHU-labelled ENA50 (122 nodes).

\begin{figure}[!h]
\centering
\vspace{-1.0em}
\includegraphics[trim=0cm 0cm 0cm 0cm, clip, width=\linewidth]{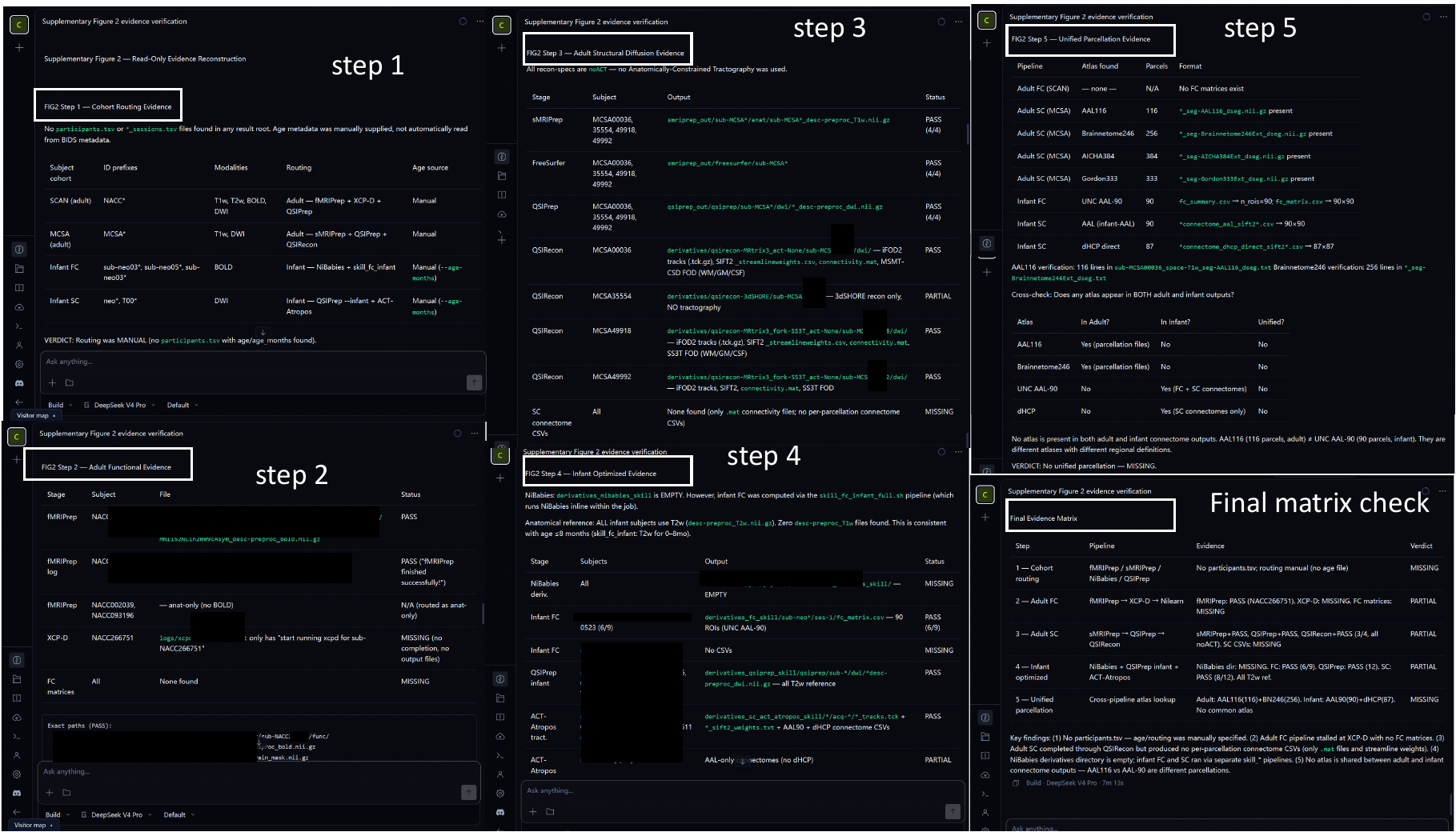}
\vspace{-1.9em}
\caption{\textbf{How \texttt{neuroimage-pre-skill} works.} }
\label{fig:supp-preproc-steps}
\end{figure}

\textbf{Step-by-step workflow.} As detailed in Fig.~\ref{supp:fig:preproc} and Fig.~\ref{fig:supp-preproc-steps}, following BIDS standardization, the \texttt{neuroimage-pre-skill} orchestrates a comprehensive preprocessing pipeline. It routes input data (T1w/T2w, fMRI, DWI) into dedicated adult or infant pathways to generate standardized functional connectivity (FC) and structural connectivity (SC) matrices:

\begin{itemize}
    \item \textbf{Step 1: Cohort Routing.} The agent automatically inspects the demographic metadata and routes the dataset into either the adult or infant processing stream to ensure age-appropriate algorithms and templates are applied.
    \item \textbf{Step 2: Adult Functional Pipeline.} For adult fMRI data, the agent chains \texttt{fMRIPrep} (for core preprocessing such as susceptibility distortion correction and co-registration) with \texttt{XCP-D} (for rigorous denoising, nuisance regression, and motion censoring). Finally, \texttt{Nilearn} is utilized to extract FC matrices using shared parcellation schemes.
    \item \textbf{Step 3: Adult Structural \& Diffusion Pipeline.} Anatomical data is processed via \texttt{sMRIPrep} (tissue segmentation, spatial normalization) and \texttt{FreeSurfer} (surface reconstruction). Concurrently, DWI data is fed into \texttt{QSIPrep} for denoising and eddy correction, followed by \texttt{QSIRecon} (utilizing MSMT-CSD and ACT with SIFT2) to generate precise SC matrices.
    \item \textbf{Step 4: Infant-Optimized Pipeline.} For infant cohorts, the agent adapts to age-specific contrast differences (e.g., relying on T2w for 0-8 months and T1w for >9 months). It leverages \texttt{NiBabies} for specialized anatomical and functional preprocessing, and employs \texttt{ACT-Atropos} paired with \texttt{SS3T-CSD} to ensure accurate infant structural tractography and SC generation.
    \item \textbf{Step 5: Unified Parcellation.} Regardless of the cohort pathway, both pipelines converge to produce outputs mapped to common, shared parcellation atlases (e.g., AAL116, Brainnetome246), ensuring downstream compatibility for multi-cohort analyses.
\end{itemize}





\subsubsection{ Quality control (\texttt{qc-agent-skill})}
\label{sec:qc}

\begin{figure}[h]
\centering
\includegraphics[trim=0cm 0cm 0cm 0cm, clip, width=0.99\linewidth]{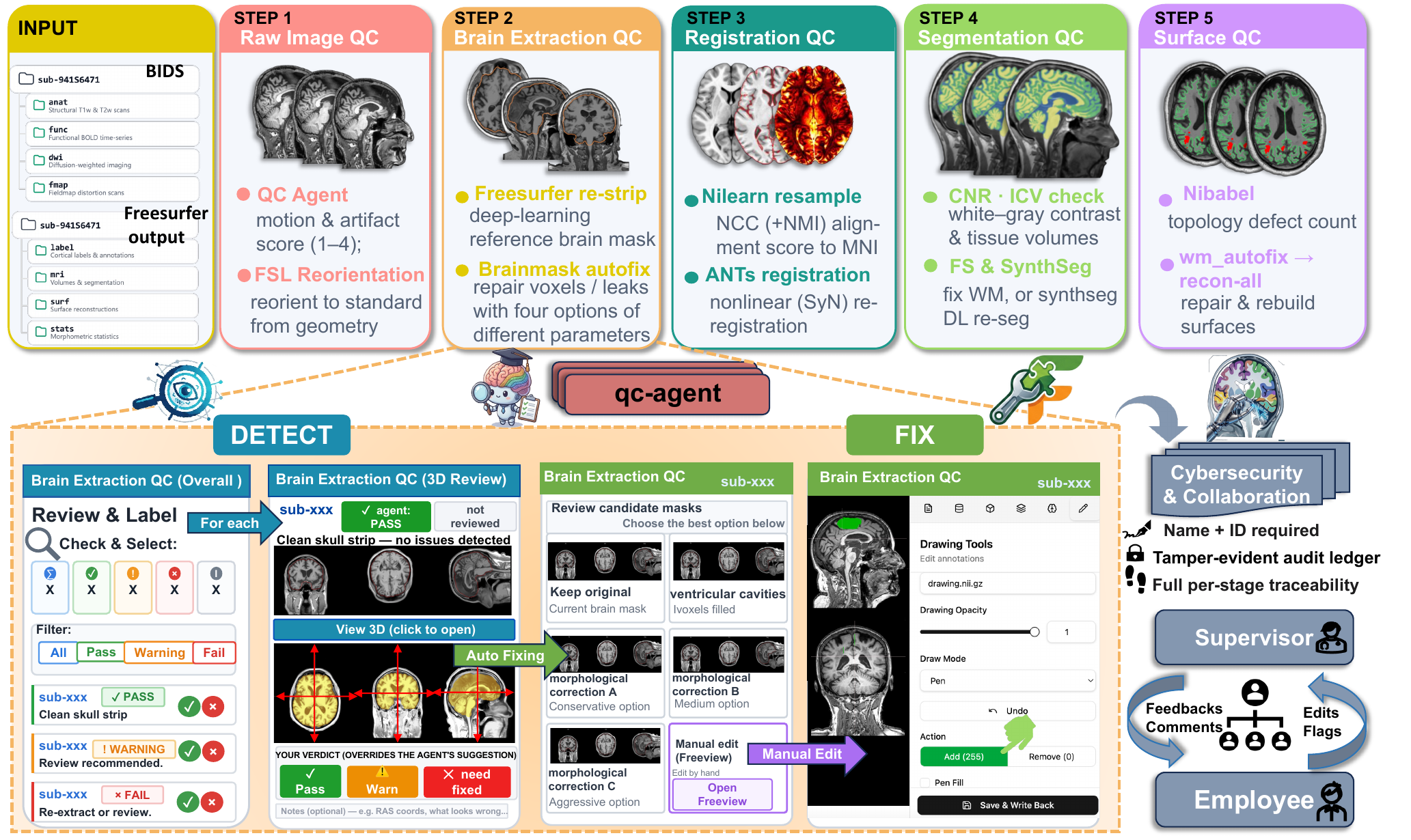}
\caption{\textbf{Structural MRI quality control -- \texttt{qc-agent-skill}.} A five-step review of T1w data runs one loop at each stage: detect and localize a defect with the skill's own scripts, show an annotated image, propose a fix, get approval, apply the fix with a delegated tool, and show a before/after check. Detection, correction, and viewing stay on separate tools. Source data is read-only, and a governance layer (reviewer identity, append-only ledger, supervisor/reviewer workflow) ensures strict management oversight and makes every decision traceable.}
\label{fig:qcflow}
\end{figure}

As shown in Fig. \ref{fig:qcflow}, the QC stage is where the human-in-the-loop model shows most plainly. It runs five structural checks on T1w data: raw image, brain extraction, registration, segmentation, surface. 

\textbf{Rule 1: Detection and fixing stay on separate tools.}
The skill's own scripts detect and localize a problem. Fixes go to the tools built for them: FreeSurfer editing (\texttt{brainmask\_autofix}, \texttt{wm\_autofix}, guided manual edits) and ANTs. Viewing uses an embedded 3-D viewer. The QC agent never edits a mask or white-matter volume by hand, which keeps the numerically consequential edits inside the same auditable tools that downstream stages expect.

\textbf{Rule 2: Grading is cohort-relative.}
No universal anatomical thresholds exist~\citep{mriqc}, so the skill flags a subject as an outlier against its own cohort and keeps a conservative absolute floor only as a safety net. Grading then tracks the scanner, protocol, and population on its own.

\textbf{Output target and audit workflow.}
Before any check, the agent fixes where corrected and kept data goes: \emph{overwrite} in place, \emph{copy} kept-and-fixed subjects to a new folder, or a \emph{custom} path. The source is never mutated silently. A \emph{dismiss} action drops a subject into one shared list that every later step filters, so it never returns downstream. A \emph{sign-off} modal gates every export: it asks for the reviewer's name and institutional ID plus a confirmation, and it appends a row to an append-only ledger (\texttt{\{by, onyen, step, at, summary\}}, where \texttt{onyen} is the institutional username), so a supervisor can see who confirmed each batch. A closing report lists every dismissed subject and deletes nothing.

\textbf{Five checkpoints and their metrics.}
\underline{\emph{(1) Raw image}}: motion and artifact scoring on the IBIS/NIRAL 1--4 scale, ghost-ratio, coverage/FOV, and intensity clipping give a stricter-than-MRIQC PASS/WARN/FAIL grade, and an orientation fingerprint plus \texttt{fslreorient2std} standardizes orientation. Only empty and wrong-modality scans stop the run. \underline{\emph{(2) Brain extraction}}: a hole count flagged by the Iglewicz--Hoaglin modified~$Z$ ($0.6745\cdot(x-\mathrm{median})/\mathrm{MAD}$) marks a cohort outlier, and over-tightness is measured against SynthStrip~\citep{synthstrip}, the reviewer picks from a watershed vertex-sweep ladder (H, A, B, C) or paints a correction. \underline{\emph{(3) Registration}}: normalized cross-correlation against the MNI template, flagged at cohort mean $-\,2\sigma$ with a floor of normalized cross-correlation $<0.40$, and TalAviQA $\ge 0.96$ for FreeSurfer outputs, confirmed outliers get ANTs full SyN. \underline{\emph{(4) Segmentation}} (T1w): MRIQC-style WM--GM contrast-to-noise and intracranial-volume tissue fractions grade Pass/Warning/Fail, \texttt{wm\_autofix} plus a partial \texttt{recon-all} repairs fragmentation, and more than 50\% missing labels fall back to SynthSeg~\citep{synthseg}. \underline{\emph{(5) Surface}} (T1w): monitors the topology-defect count, defined as the total number of surface vertices corrected by FreeSurfer's topology fixer ($\Sigma$ non-zero \texttt{defect\_labels}, lh+rh)~\citep{euler}. It flags outliers exceeding the cohort median $+\,k\cdot\mathrm{MAD}$ ($k\approx3$). For repairs, combining \texttt{wm\_autofix} with a targeted partial \texttt{recon-all} rebuilds the surface in just 2--4~h, compared to 8--12~h for a full rerun.

\textbf{Scalable cohort review and human-agent teaming.} Each QC check generates a self-contained HTML dashboard, securely exposed via local ports or SSH tunnels, featuring subject cards with three-view screenshots and an interactive, WebGL-optimized 3-D viewer (\texttt{niivue}). Within this interface, the agent automates the mechanism (computing metrics, localizing defects, and proposing fixes), while the human retains full judgment authority (pass/fail/dismiss and fix selection). Keyboard shortcuts accelerate high-throughput labeling, and hard cases follow a multi-tier escalation path from agent flags to reviewer overrides, and ultimately to supervisor inspection. Crucially, every decision is immutably recorded via JSON in a centralized ledger, ensuring robust management oversight, preserving institutional memory, and guaranteeing the provenance of all quality control actions.

\begin{figure}[!h]
\centering
\includegraphics[trim=0cm 0cm 0cm 0cm, clip, width=\linewidth]{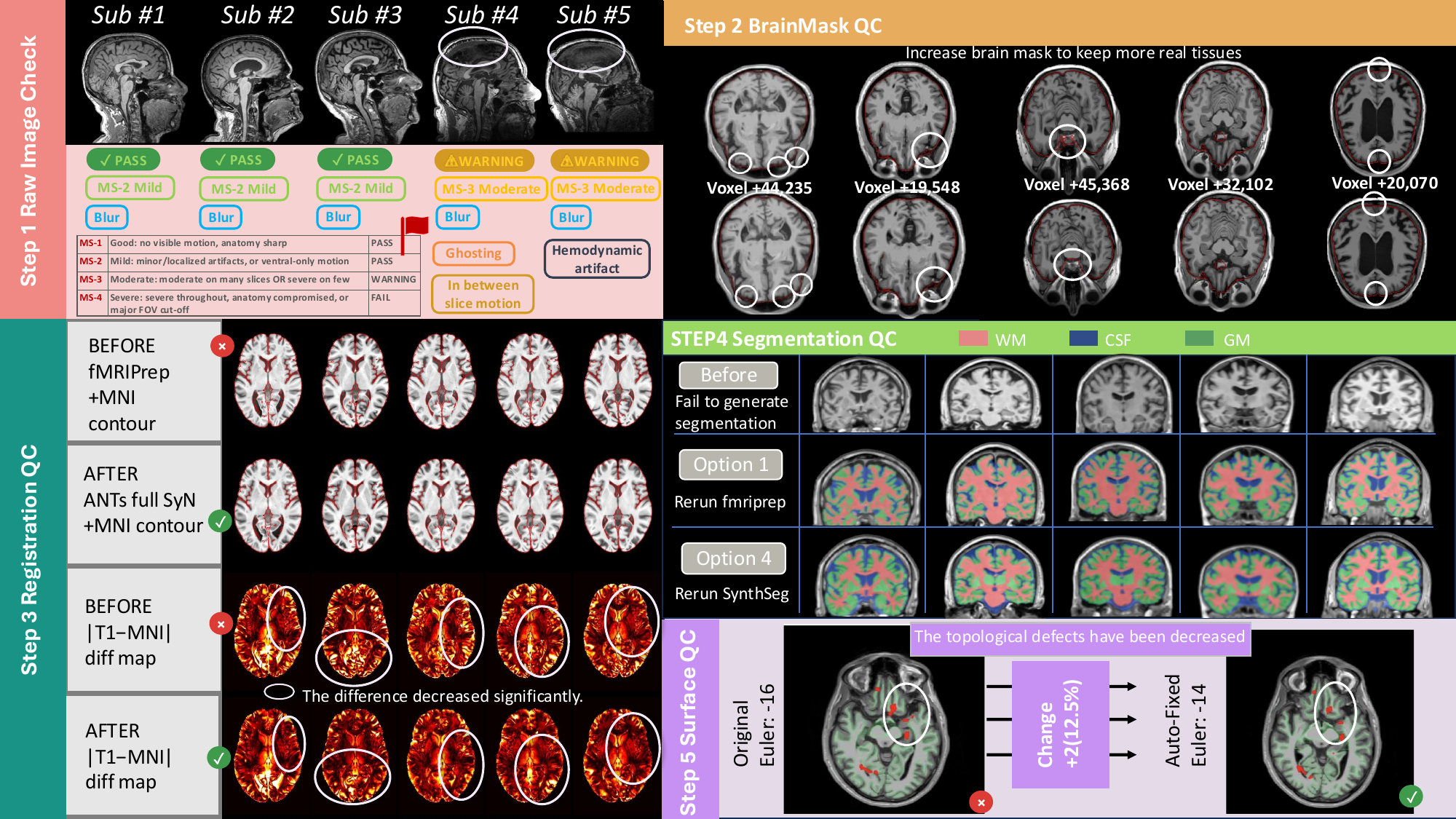}
\caption{\textbf{QC Agent in action.} Demonstration of the detect--propose--approve--fix--verify loop across five structural QC steps: \textbf{(1)}~raw-image motion grading; \textbf{(2)}~brain-mask expansion; \textbf{(3)}~registration repair; \textbf{(4)}~segmentation restoration; and \textbf{(5)}~surface topology-defect reduction. Detection, correction, and visualization are decoupled to preserve read-only source data and maintain an auditable decision ledger.}
\label{fig:qc-result}
\end{figure}
\begin{figure}[!h]
\centering
\vspace{-1.2em}
\includegraphics[trim=0cm 0cm 0cm 0.0cm, clip, width=\linewidth]{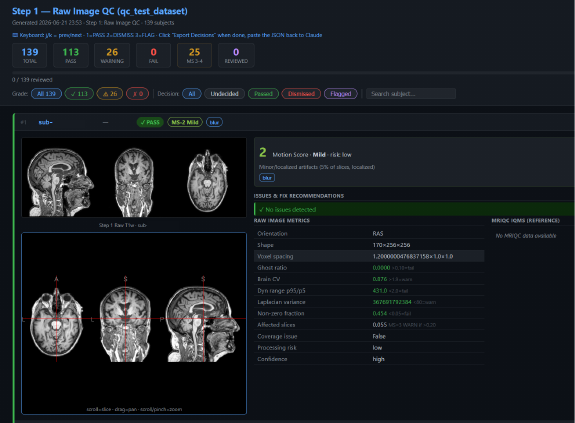}
\vspace{-1.5em}
\caption{\textbf{How \texttt{qc-agent-skill} works -- Step 1.} }
\label{fig:supp-qc-step1}
\vspace{-1.5em}
\end{figure}
\begin{figure}[!h]
\centering
\vspace{-1.5em}
\includegraphics[trim=0cm 0cm 0cm 0.0cm, clip, width=\linewidth]{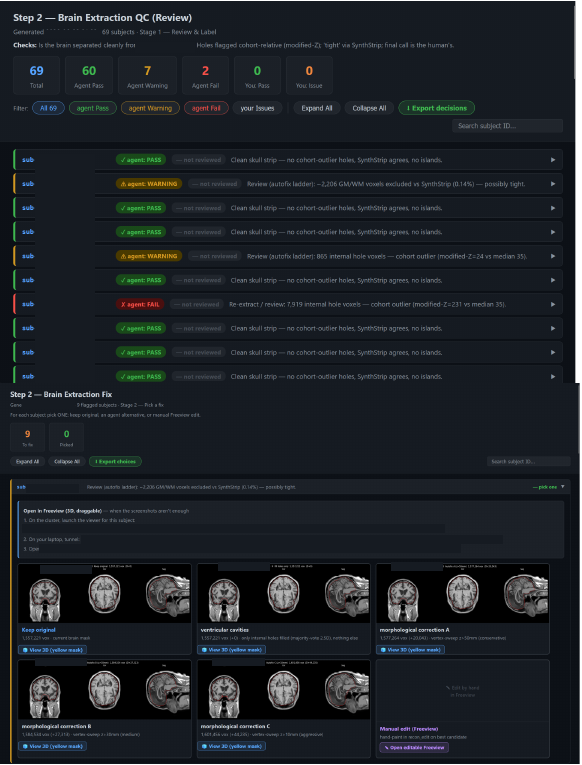}
\vspace{-1.5em}
\caption{\textbf{How \texttt{qc-agent-skill} works -- Step 2.} }
\label{fig:supp-qc-step2}
\end{figure}
\begin{figure}[!h]
\centering
\vspace{-1.5em}
\includegraphics[trim=0cm 0cm 0cm 0.0cm, clip, width=\linewidth]{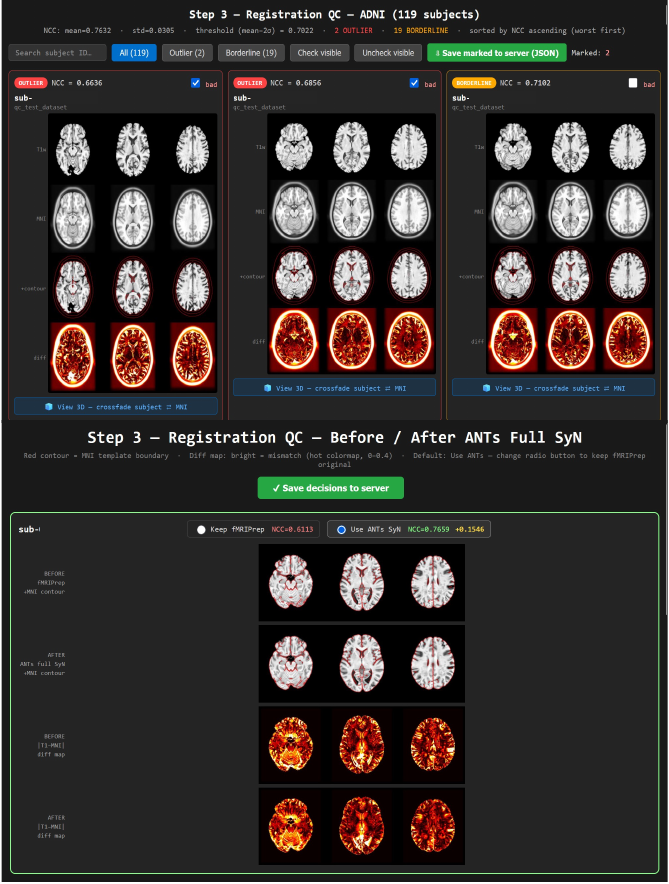}
\vspace{-1.5em}
\caption{\textbf{How \texttt{qc-agent-skill} works -- Step 3.} }
\label{fig:supp-qc-step3}
\end{figure}
\begin{figure}[!h]
\centering
\vspace{-1.5em}
\includegraphics[trim=0cm 0cm 0cm 0.0cm, clip, width=\linewidth]{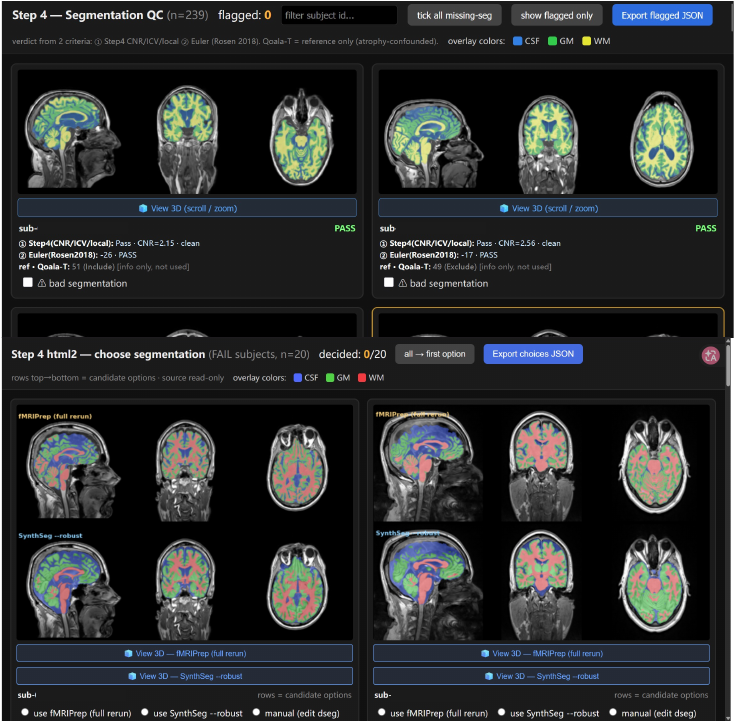}
\vspace{-1.5em}
\caption{\textbf{How \texttt{qc-agent-skill} works -- Step 4.} }
\label{fig:supp-qc-step4}
\end{figure}
\begin{figure}[!h]
\centering
\vspace{-2.5em}
\includegraphics[trim=0cm 0cm 0cm 0.0cm, clip, width=0.7\linewidth]{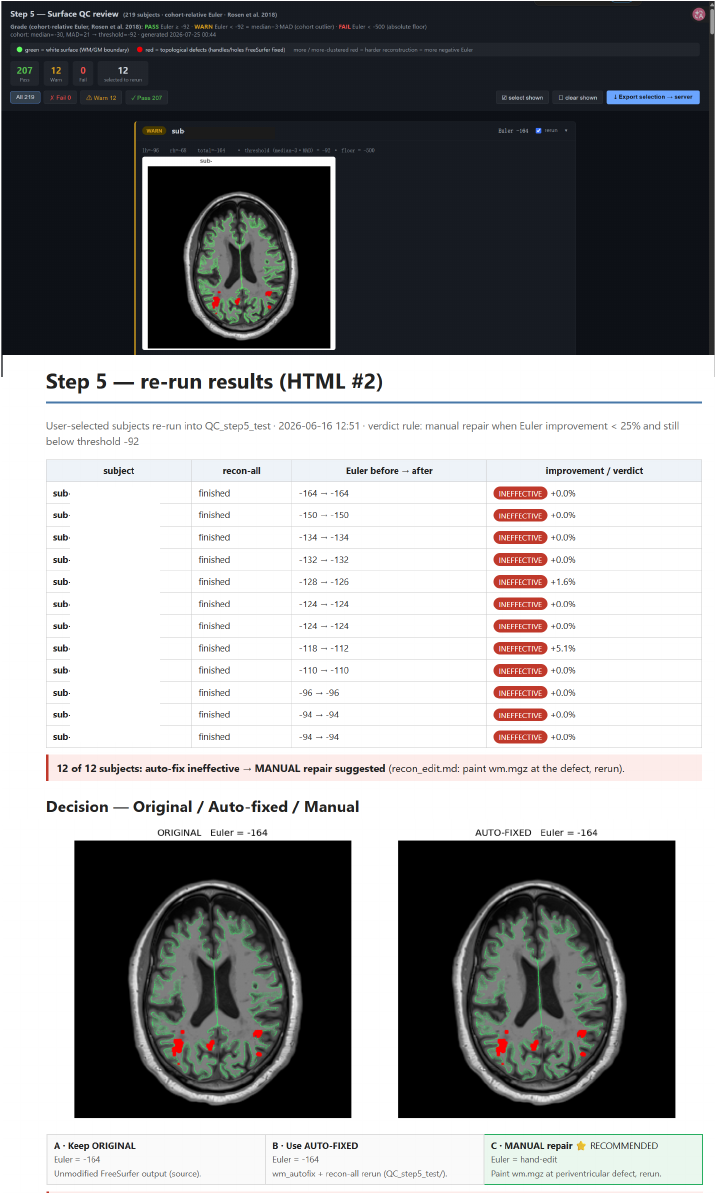}
\vspace{-1.0em}
\caption{\textbf{How \texttt{qc-agent-skill} works -- Step 5.} }
\label{fig:supp-qc-step5}
\end{figure}

\textbf{Step-by-step workflow.} Fig.~\ref{fig:qc-result} shows the five QC steps on production subjects. As demonstrated there, the QC Agent employs a decoupling strategy (separating detection, correction, and visualization) governed by a strict detect--propose--approve--fix--verify loop. The specific five-step structural QC process is as follows (Figs.~\ref{fig:supp-qc-step1}--\ref{fig:supp-qc-step5}):

\begin{itemize}
    \item \textbf{Step 1: Raw Image QC.} The agent evaluates the raw images for motion and artifacts, assigning a severity score (1–4). If necessary, FSL Reorientation is proposed to reorient the image to standard geometry.
    \item \textbf{Step 2: Brain Extraction QC.} The agent compares the current brain mask against a deep-learning reference mask. If leaks or over-trimming are detected, it proposes morphological autofix options (Conservative, Medium, Aggressive) or allows for manual editing via Freeview.
    \item \textbf{Step 3: Registration QC.} Nilearn is used to compute the Normalized Cross-Correlation (NCC) and Normalized Mutual Information (NMI) alignment scores to the MNI template. Poor alignments trigger a proposed fix using ANTs nonlinear (SyN) re-registration.
    \item \textbf{Step 4: Segmentation QC.} The agent checks the Contrast-to-Noise Ratio (CNR) and Intracranial Volume (ICV) for white-gray contrast and tissue volumes. Fixes involve FreeSurfer White Matter (WM) edits or SynthSeg deep-learning re-segmentation.
    \item \textbf{Step 5: Surface QC.} Using Nibabel, the agent counts topological defects (Euler number). If the defect count is abnormally high, it executes \texttt{wm\_autofix} and \texttt{recon-all} to repair and rebuild the cortical surfaces.
\end{itemize}

\section{Experiments and Results}
\label{sec:experiments}
We structure experiments around three modules: validate data standardization via \texttt{dcm2bids-skill}, evaluate modality-specific preprocessing (\texttt{neuroimage-pre-skill}) on infant and adult cohorts, and assess \texttt{qc-agent-skill}-driven automated detection and repair.


\subsection{Evaluation on {\texttt{dcm2bids-skill}}}

\begin{figure}[h]
\centering
\includegraphics[width=1\linewidth, height=0.48\textheight, keepaspectratio]{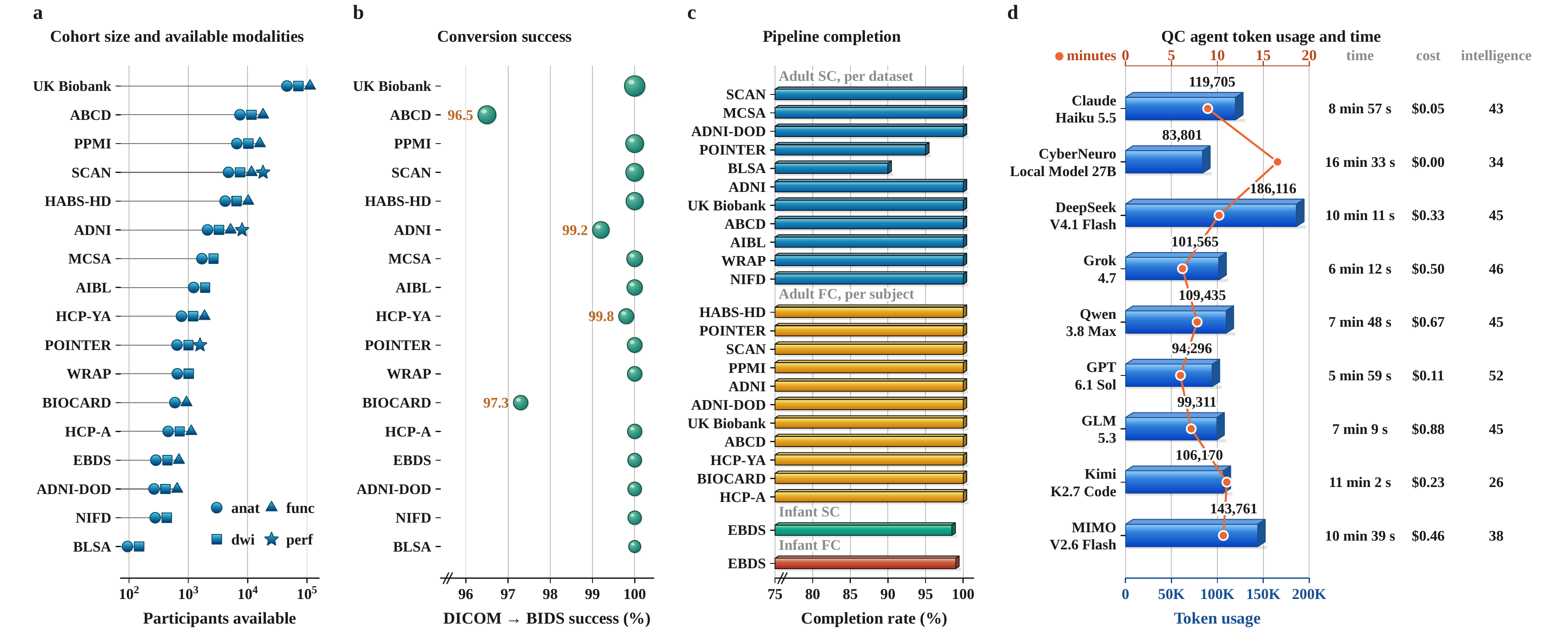}
\vspace{-1.0em}
\caption{\textbf{Validation results.} \textbf{(a)}~The 17 cohorts and their
modality coverage. \textbf{(b)}~DICOM-to-BIDS conversion completeness.
\textbf{(c)}~Neuroimage preprocessing pipeline completion, including adult and infant.
\textbf{(d)}~\texttt{qc-agent-skill}'s token usage, cost, and timing on 9 LLM models with different intelligence score by \href{https://artificialanalysis.ai/}{Artificial Analysis}. \emph{Note:} All inputs are raw and unfiltered.}
\label{fig:result}
\end{figure}

We evaluated the skill on the 17 testbed cohorts (Fig.~\ref{fig:result}(a)) and report the conversion success rate of each in Fig.~\ref{fig:result}(b). The rate was between 99.2\% and 99.8\% for most cohorts, and the lowest was 96.5\%. We traced every conversion failure to problems in the source DICOM data (malformed headers, missing phase-encoding metadata, or incomplete series), none originated in the conversion tool itself. All successfully converted data also passed BIDS validation and were classified correctly, so classification accuracy (Fig.~\ref{fig:dcmbids2}) matched the conversion success rates. Up to classification, conversion was therefore the only step that limited yield, and in these cohorts a correct BIDS structure was sufficient for correct modality routing.

\subsection{Evaluation on \texttt{neuroimage-pre-skill}}
\label{sec:preskill-eval}
We evaluated the adult pipeline on 16 cohorts and the infant pipeline on \href{https://www.med.unc.edu/psych/research/programs/early-brain-development-research/}{UNC-EBDS}.

\textbf{Adult connectivity and surface reconstruction.}
\label{sec:adult}
As shown in Fig.~\ref{fig:result}(c), the adult preprocessing pipeline achieved robust completion rates for both SC and FC workflows across the 16 evaluated adult cohorts. Importantly, every pipeline failure was traced back to inherent defects in the source data rather than processing fragility.
\textit{Structural and surface outcomes:} FreeSurfer surface reconstruction successfully completed for all 11 cohorts that finished processing. Across these datasets, macro-anatomical metrics (e.g., intracranial volume, cortical thickness, and brain-segmentation volume) fell within biologically plausible ranges for 89--100\% of subjects. For SC, we evaluated {10{,}807} completed sessions across 11 cohorts, with 99.9\% of the generated matrices passing every rigorous anatomical check.
\textit{Functional connectivity and reliability:} We assessed {48{,}293} completed resting-state BOLD runs from 11 cohorts {7{,}755} participants), confirming every resulting matrix was mathematically valid. After stringent motion scrubbing, {2{,}076} runs were flagged for retaining insufficient data ($<240$\,s). This filtering successfully isolated edge cases, such as the 10-volume secondary runs in PPMI that caused near-zero two-run correlations during pilot testing. Among the surviving runs, 99.9\% passed all anatomical criteria (Table~\ref{tab:adult}).

\begin{table}[!h]
\centering
\caption{{Plausibility checks on adult connectomes.} Counts are completed matrices passing each check. SC: ADNI, ADNI-DOD, MCSA, POINTER, SCAN, ABCD, HCP-YA (10{,}807 sessions, 10{,}798 participants). FC: resting-state runs from ADNI, PPMI, ABCD (48{,}293 runs, 7{,}755 participants).}
\label{tab:adult}
\small
\setlength{\tabcolsep}{4pt}
\resizebox{\linewidth}{!}{
\begin{tabular}{l l r r}
\toprule
\textbf{Check} & \textbf{Criterion} & \textbf{SC} & \textbf{FC} \\
\midrule
Validity               & symmetric, finite; FC: unit diagonal, $|r| \le 1$           & 10{,}807 & 48{,}293 \\
Coverage               & empty parcels $\le 5\%$                                      & 10{,}803 & 48{,}235 \\
Data sufficiency$^{a}$ & SC: complete gradient table; FC: $\ge 240$~s after scrubbing & 10{,}807 & 46{,}217 \\
Distance decay         & $\rho(\text{connection}, \text{distance}) < 0$               & 10{,}807 & 48{,}290 \\
Hemisphere             & intra- $>$ heterotopic inter-hemispheric                     & 10{,}807 & 48{,}286 \\
Homotopy$^{b}$         & homotopic $>$ heterotopic                                    & 10{,}804 & 48{,}289 \\
Cohort similarity$^{c}$ & $\rho$ to leave-one-out mean $\ge$ median $-$ 3\,MAD        & 10{,}142 & 47{,}806 \\
\midrule
\textbf{All checks}$^{d}$             &                                       & \textbf{10{,}142 (93.8\%)} & \textbf{45{,}956 (95.2\%)} \\
\textbf{Anatomically plausible}$^{e}$ &                                       & \textbf{10{,}801 (99.9\%)} & \textbf{46{,}176 (99.9\%)} \\
\bottomrule
\end{tabular}}
\end{table}

\textbf{Infant connectivity.}
\label{sec:infant}
We applied the infant pipeline into \href{https://www.med.unc.edu/psych/research/programs/early-brain-development-research/}{UNC-EBDS} dataset (age $<$ 1y). To ensure pipeline stability, early input defects (e.g., missing phase-encoding metadata, truncated DWIs, and missing gradient files) were either repaired during BIDS conversion or screened out prior to submission (Table~\ref{supp:tab:infantfail}). As a result, both the SC and FC routes successfully generated connectivity matrices across all age-appropriate atlases for every submitted session. We validated the integrity and reliability of each AAL-90 matrix using the topological and statistical criteria outlined in Table~\ref{tab:infant}. 
\textit{For the SC route}, which processed 224 sessions (223 infants) containing both T2w and usable DWI data, 91\% of sessions were ultimately usable. Rejections fell into three distinct categories: (1) DWI-to-T2w misregistration on the Allegra scanner (which bypassed the pipeline's matrix-level QC) (2) template-registration failures (which the pipeline successfully flagged) and (3) statistical outliers in homotopic connectivity or cohort similarity. 
\textit{For the FC route}, processing 311 sessions (310 infants 309 neonatal and two aged 6--7 months) with structural and resting BOLD images, 98.4\% of sessions passed every check. All FC failures were exclusively due to insufficient BOLD data retention (<240s post-scrubbing). Furthermore, valid FC matrices exhibited robust biological plausibility, with homotopic connectivity proving far stronger than heterotopic connectivity (median 0.53 versus 0.04).

\begin{table}[!h]
\centering
\caption{The twelve infant SC non-completions. Each is a defect in the input DWI, not a pipeline error.}
\label{supp:tab:infantfail}
\small
\setlength{\tabcolsep}{3pt}
\begin{tabular}{l c r}
\toprule
\textbf{Failure cause} & \textbf{Pipeline fault?} & \textbf{$n$} \\
\midrule
DWI missing JSON (no phase-encoding metadata) & No (data) & 8 \\
DWI truncated (bval count $\neq$ volumes)      & No (data) & 2 \\
Missing bval/bvec                              & No (data) & 1 \\
Corrupt multi-run DWI                          & No (data) & 1 \\
\midrule
\textbf{Total}                                 &           & \textbf{12} \\
\bottomrule
\end{tabular}
\end{table}

\begin{table}[!h]
\centering
\caption{{Plausibility checks on infant connectomes} (EBDS, AAL-90). Counts are sessions passing each check.}
\label{tab:infant}
\small
\setlength{\tabcolsep}{4pt}
\resizebox{\linewidth}{!}{
\begin{tabular}{l l r r}
\toprule
\textbf{Check} & \textbf{Criterion} & \textbf{SC} & \textbf{FC} \\
\midrule
Validity          & symmetric, finite; FC: unit diagonal, $|r| \le 1$              & 224 & 311 \\
Coverage          & empty parcels $\le 5\%$                                         & 220 & 311 \\
Data sufficiency$^{a}$ & SC: complete gradient table, $\ge 12$ directions; FC: $\ge 240$~s of BOLD & 224 & 306 \\
Distance decay    & $\rho(\text{connection}, \text{distance}) < 0$                  & 224 & 311 \\
Hemisphere        & intra- $>$ heterotopic inter-hemispheric                        & 224 & 311 \\
Homotopy$^{b}$    & homotopic $>$ heterotopic                                       & 206 & 311 \\
Cohort similarity & $\rho$ to leave-one-out mean $\ge$ median $-$ 3\,MAD            & 188 & 311 \\
\midrule
\textbf{All checks}$^{c}$ &                                                         & \textbf{188} & \textbf{306} \\
\textbf{Usable}$^{d}$     &                                                         & \textbf{204 (91\%)} & \textbf{306 (98.4\%)} \\
\midrule
Sessions (infants)  &                                                               & 224 (223) & 311 (310) \\
\bottomrule
\end{tabular}}

\vspace{2pt}
\begin{minipage}{\linewidth}
\footnotesize
$^{a}$~SC is screened before processing, since tractography has no scan-length requirement but needs a complete diffusion gradient table; FC is checked after motion scrubbing, since too few retained volumes make correlations unreliable.
$^{b}$~SC additionally requires at least 50\% of homotopic edges to be present; FC matrices are dense.
$^{c}$~Passes every row above.
$^{d}$~Passes every row, or fails only cohort similarity with $\rho \ge 0.85$ (16 SC sessions); usable FC equals all checks.
\end{minipage}
\end{table}

\begin{figure}[!h]
\centering
\includegraphics[width=\linewidth]{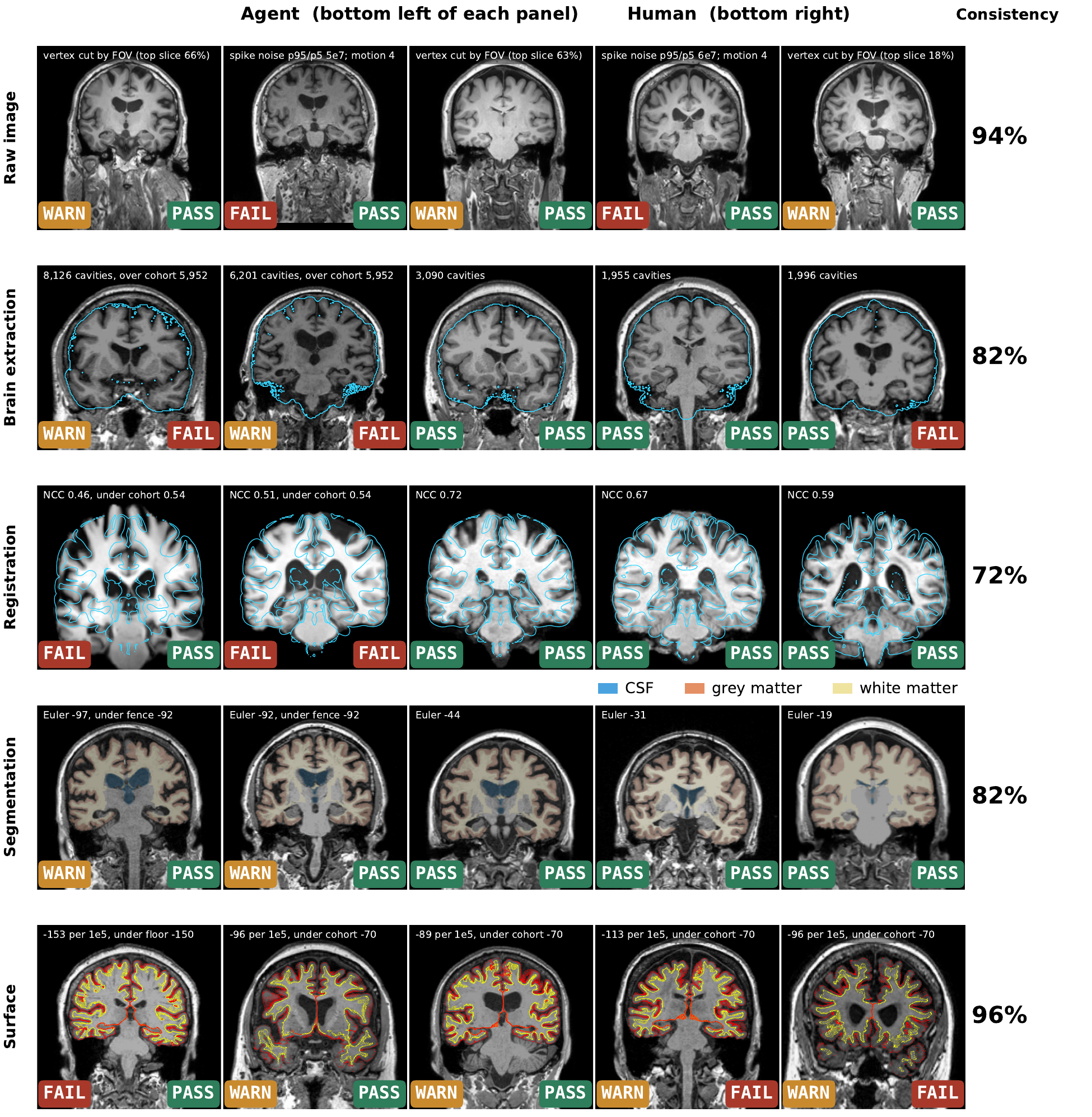}
\vspace{-1.5em}
\caption{\textbf{QC consistency between the agent and human expert on the ADNI dataset.} Each panel displays the specific metric that triggered the agent's decision. The agent's decision is shown at the bottom left, and the human rater's decision is at the bottom right. The percentage in the right panel indicates the overall consistency rate between the two across ADNI subjects. A \textsc{warn} decision indicates that the case requires manual review and a final decision by a human expert.}
\label{fig:agent-vs-human}
\end{figure}

\subsection{Evaluation of the \texttt{qc-agent-skill}}
\label{sec:qc-results}
We evaluated the \texttt{qc-agent-skill} across three dimensions: defect detection validity, decision reliability, and operational scalability (cost and time).

\textbf{Validity and sensitivity.} To establish ground truth, we conducted an error-injection experiment on 500 ADNI subjects (100 intact controls, 400 with injected defects at varying severities. See Table ~\ref{tab:injection-design}). The agent achieved decent specificity (all intact subjects passed) and consistently detected defects with higher sensitivity at \textsc{fail} severities than \textsc{warn} severities across all steps (See table ~\ref{tab:injection-detection}). Beyond synthetic errors, we verified that the agent's flags track real-world anatomical degradation. For instance, the agent's registration grades strongly correlated with white-matter tissue overlap computed by MRIQC ($\rho = 0.80$, $p < 10^{-11}$), and surface defect density correlated with the coefficient of joint variation ($\rho = -0.71$). Importantly, these flags meaningfully impact downstream analysis: subjects flagged at registration or segmentation exhibited significantly lower mean cortical thickness ($d = -1.02$ and $d = -0.78$, respectively $p < 0.05$), proving the flags successfully isolate compromised morphometry data (Table~\ref{tab:e3}). To evaluate the consistency between the agent and human judgment, we compared the agent's calls against a human expert's manual QC on 50 ADNI reconstructions. As shown in Fig.~\ref{fig:agent-vs-human}, the agent achieved a high agreement rate with the human expert, reaching 80-90\%. Notably, in discordant cases where the agent issued a \textsc{fail} decision but the human rater initially marked a \textsc{pass}, a subsequent cross-validation by three independent experts confirmed that the agent's decision was actually correct. This reveals that manual QC is inherently susceptible to subjective bias, and that visually inspecting all slices in complex 3-D neuroimages is prone to human oversight, strongly underscoring the practical value of our QC-Agent. To the best of our knowledge, this is the \textbf{first work} to propose an semi-autonomous agent for neuroimaging QC.

\begin{table}[!h]
\centering
\footnotesize
\setlength{\tabcolsep}{3pt}
\caption{Error-injection design. One injected defect per case, at one of two
severity levels, anchored in physical units rather than in the agent's own metric.
All five steps are run for every case, so each defect yields a five-element
detection vector rather than a single ``first flagged at'' step.}
\label{tab:injection-design}
\begin{tabular}{@{}lrll@{\hskip 14pt}ll@{}}
\toprule
 & & & \multicolumn{2}{c}{Injected severity} & \\
\cmidrule(lr){4-5}
Step & \# & Injection angle & WARN & FAIL & Detection basis \\
\midrule
\#1
  & 1  & Motion artefact           & $<$\,10\,\% slices  & $>$\,30\,\% slices   & Ghost ratio \\
  & 2  & FOV truncation            & 5--8\,mm cut        & 20\,mm cut           & Coverage check \\
  & 3  & Subject swap              & within-cohort       & cross-cohort         & Sidecar / ID check \\
\midrule
\#2
  & 4  & Missing voxels            & 0.3\,\% of brain    & 2\,\% of brain       & Cohort modified $Z$ \\
  & 5  & Leakage                   & $+$2\,\% dura       & $+$10\,\% scalp      & SynthStrip $\Delta$ \\
  & 6  & Directed under-inclusion  & 1--2\,\% GM/WM      & 8--10\,\% GM/WM      & SynthStrip $\Delta$ (local) \\
\midrule
\#3
  & 7  & Rigid misalignment        & 3\,mm; $3^{\circ}$  & 10\,mm; $12^{\circ}$ & NCC vs.\ MNI \\
  & 8  & Local nonlinear displ.    & 4\,mm, regional     & 10\,mm, regional     & Regional diff.\ map \\
  & 9  & L/R flip                  & header only         & data flipped         & L/R header check \\
\midrule
\#4
  & 10 & Missing label             & 20\,\% CSF          & class zeroed         & Missing frac.; ICV \\
  & 11 & Boundary shift            & 1 voxel             & 3 voxels             & ICV drift; CNR \\
  & 12 & dseg--T1 shift            & 2\,mm               & 6\,mm                & T1--dseg overlay \\
\midrule
\#5
  & 13 & WM punch-through          & 20 sites            & 150 sites            & Euler number \\
  & 14 & Pial leakage              & small dura          & large dura           & Pial--T1 overlay \\
  & 15 & Vertex perturbation       & 1\,mm               & 3\,mm                & Surface outlines \\
\bottomrule
\end{tabular}

\vspace{3pt}
\parbox{\textwidth}{\footnotesize\raggedright
\emph{Targets}: raw T1w (1--3); \texttt{brainmask.mgz} (4--6); MNI-space T1 and transform
(7--9); \texttt{dseg} (10--12); \texttt{wm.mgz} and \texttt{surf} (13--15). Rows A and B are not graded by severity: A measures whether the
failure is reported or crashes the run, B supplies the false-positive rate.}
\end{table}

\begin{table}[!h]
\centering
\caption{{Injected-defect detection by QC step.} Counts are the share of
injected defects flagged at that step, at each injected severity level;
$\Delta$ is FAIL $-$ WARN in percentage points. 400 injected cases (one defect
each; 15 injection angles $\times$ 2 severity levels, Table~\ref{tab:injection-design})
and 100 intact cases, drawn as ten batches of 50. All five steps were run for
every case irrespective of early flags; per-step rates are mixtures over the
three injection angles assigned to that step.}
\label{tab:injection-detection}
\small
\setlength{\tabcolsep}{4pt}
\begin{tabular}{l r r r}
\toprule
\textbf{Step} & \textbf{WARN (\%)} & \textbf{FAIL (\%)} & \textbf{$\Delta$ (pp)} \\
\midrule
1 & 92.1 & 98.1 & $+$6.0 \\
2 & 87.8 & 98.5 & $+$10.7 \\
3 & 91.6 & 95.3 & $+$3.7 \\
4 & 97.2 & 98.1 & $+$0.9 \\
5 & 80.6 & 90.1 & $+$9.5 \\
\midrule
\textbf{Uninjected} & \multicolumn{3}{r}{\textbf{no step flagged (specificity 100\%)}} \\
\bottomrule
\end{tabular}
\end{table}

\begin{table}[!h]
\centering
\caption{The five QC steps on 600 ADNI subjects, each graded by the rule the
step actually ships with and checked against an MRIQC metric computed from the
raw image, which never sees the reconstruction the step grades.}
\label{tab:e3}
\begin{tabular}{@{}clrlr@{\,}lr@{\,}l@{}}
\toprule
\multicolumn{3}{c}{QC step}
& \multicolumn{3}{c}{Agreement with MRIQC}
& \multicolumn{2}{c}{Cortical thickness} \\
\cmidrule(lr){1-3}\cmidrule(lr){4-6}\cmidrule(lr){7-8}
No. & Graded quantity & Flagged (\%) & MRIQC metric
& \multicolumn{2}{c}{$\rho$ (Spearman)}
& \multicolumn{2}{c}{$d$ (Cohen's)} \\
\midrule
\# 1 & Artifact grade $\downarrow$ & 12.8 & QI$_2$ $\downarrow$ & $\phantom{-}0.59$ & *** & \multicolumn{2}{c}{---} \\
\# 2 & Extraction grade $\downarrow$ & 7.8 & GM SNR $\uparrow$ & $-0.28$ & *** & $-0.39$ & * \\
\# 3 & Template NCC $\uparrow$ & 2.7 & TPM overlap $\uparrow$ & $\phantom{-}0.80$ & *** & $-1.02$ & ** \\
\# 4 & Euler number $\uparrow$ & 10.5 & CNR $\uparrow$ & $\phantom{-}0.72$ & *** & $-0.78$ & *** \\
\# 5 & Euler density $\uparrow$ & 12.7 & CJV $\downarrow$ & $-0.71$ & *** & $-0.62$ & *** \\
\bottomrule
\end{tabular}

\vspace{3pt}
\begin{minipage}{\linewidth}\footnotesize
*** $p<0.001$, ** $p<0.01$, * $p<0.05$. Steps: 1 raw image, 2 brain extraction,
3 registration, 4 segmentation, 5 surface. An arrow marks which way is better: $\uparrow$ means
a higher value is better, $\downarrow$ that a lower value is better. $\rho$ is
therefore expected to be positive where a step's arrow matches its reference's
and negative where the two oppose; every row meets that expectation.
$\rho$ correlates the ranks of the two quantities; $d$ compares mean
left-hemisphere cortical thickness between flagged subjects and the rest, in
pooled standard deviations.

Graded quantities: artifact grade and extraction grade, the verdicts steps~1
and 2 reach from several criteria at once, correlated as the ordinal grade
PASS $<$ WARN $<$ FAIL, on coverage, ghosting and intensity range for the first
and on the brain mask for the second; NCC, normalised cross-correlation to the
template; Euler number, $2-2h$ for $h$ holes in the unfixed surface; Euler
density, the same count per $10^{5}$ vertices. MRIQC metrics: QI$_2$, the
second Mortamet quality index, the fit of a noise model to the image background, which rises with structured artifact there; GM SNR, signal-to-noise ratio in grey matter; CNR, grey--white contrast-to-noise ratio;
TPM overlap, agreement of the subject's white-matter tissue probability map with the template prior; CJV, coefficient of joint variation of white- and grey-matter intensities. Step~1 detects coverage and artifact, which do not alter morphometry.
\end{minipage}
\end{table}

\textbf{Decision reliability.} Because consensus ground truth for real-world defects is elusive, we measured the agent's decision stability by running five independent QC passes on 200 ADNI subjects using default parameters. The agent demonstrated ``almost perfect'' agreement (Fleiss' $\kappa \geq 0.82$) across all five structural checks (Table~\ref{tab:reliability}). Notably, while segmentation achieved the highest raw agreement ($96.1\%$), its $\kappa$ ($0.882$) was slightly lower than registration ($\kappa = 0.901$, raw agreement $94.6\%$). This reflects the agent's robust discriminative consistency even in steps where \textsc{pass} decisions do not overwhelmingly dominate chance agreement.

\begin{table}[!h]
\centering
\small
\caption{Reliability of each QC step across repeated runs. Each step was run five
times over the same 200-subject ADNI dataset, independently and untuned.
\textit{Verdict mix}: share of runs per verdict. \textit{Agr.} ($\bar{P}$):
observed agreement; $P_e$: chance agreement from verdict shares;
$\kappa=(\bar{P}-P_e)/(1-P_e)$: Fleiss' $\kappa$, corrected for chance, not
correctness. All steps are ``almost perfect'' ($\kappa\geq0.82$).}
\label{tab:reliability}
\begin{tabular}{l l r r r r r r}
\toprule
 & & \multicolumn{3}{c}{Verdict mix (\%)} & & & \\
\cmidrule(lr){3-5}
Step & Stage & PASS & WARN & FAIL & $P_e$ & Agr.\ (\%) & $\kappa$ \\
\midrule
\#2 & Brain extraction & 72.8 & 14.9 & 12.3 & 0.568 & 92.3 & 0.822 \\
\#3 & Registration     & 61.7 & 22.7 & 15.6 & 0.457 & 94.6 & 0.901 \\
\#4 & Segmentation     & 80.6 & 10.8 &  8.6 & 0.669 & 96.1 & 0.882 \\
\#5 & Surface          & 60.8 & 29.4 &  9.8 & 0.466 & 90.4 & 0.820 \\
\bottomrule
\end{tabular}
\end{table}

\textbf{Scalability: Human effort and model-agnostic cost.} The agent radically reduces operator burden. Performing the five-step QC manually required 177 minutes per subject (including inspection and tuning), whereas the automated screening took only 25.4 seconds, i.e., a $\bf 418\times$ speedup that eliminates linear scaling costs for large cohorts (see table ~\ref{tab:e6}). Furthermore, the framework is model-agnostic. We benchmarked the detection pass across nine frontier LLMs using an identical prompt. All models successfully completed the task (Fig.~\ref{fig:result}(d)), with mean token usage of 116,018 and an average API cost of \$0.36 per subject. For maximum privacy and zero marginal API cost, we successfully deployed a local Wanda model in \href{https://wanda-cyberbench.com/}{CyberNeuro} \citep{ren2026cyberneuro}, maintaining feasible execution times without transmitting PHI off-site.

\begin{table}[!h]
\centering
\caption{Human effort for the five QC steps; every figure in the table is per
subject. Inspection and tuning are means over the ADNI dataset for the manual arm, automated screening likewise for the agent.  Compute is shown for reference and is not part of the comparison, since the same jobs are submitted in either arm. It
is also the one column that does not accumulate: subjects are independent, so the
cluster runs them concurrently and a cohort's wall clock is set by how many jobs
run at once rather than by the number of subjects, whereas one operator's
inspection and tuning add up subject by subject. Script writing is paid once per
cohort and is listed separately for that reason.}
\label{tab:e6}
\begin{tabular}{@{}lrrrrr@{}}
\toprule
& \multicolumn{3}{c}{Manual (min)} & Agent & Compute \\
\cmidrule(lr){2-4}
QC step & Inspect & Tune & Total & \textbf{(s)} & \textbf{(min)} \\
\midrule
\# 1~~Raw image & 22 & --- & 22 & 4.8 & --- \\
\# 2~~Brain extraction & 17 & 17 & 34 & 12.2 & 1.9 \\
\# 3~~Registration & 15 & 15 & 30 & 4.0 & 7.7 \\
\# 4~~Segmentation & 19 & 20 & 39 & 2.5 & 2.7 \\
\# 5~~Surface & 25 & 27 & 52 & 1.9 & 102 \\
\midrule
All five steps & 98 & 79 & \textbf{177} & \textbf{25.4} & 114 \\
Script writing, once & \multicolumn{3}{r}{160} & --- & --- \\
\bottomrule
\end{tabular}
\end{table}

\subsection{Effectiveness of the agent-based architecture}
\label{sec:ablation}

To test whether the framework's performance comes from the agent's decisions or from the pipeline orchestration around them, we replaced the agent layer with a static baseline built from hard-coded keyword and header rules, keeping the rest of the pipeline unchanged. We compared the two on the following three tasks (Table~\ref{tab:e1}).

\begin{table}[!h]
\centering
\caption{Ablating the agent layer against a fixed rule baseline. The three arms answer different questions and are scored in different units, so they are reported separately. Classification is scored against expert-adjudicated series labels; routing against the policy applied to processable modalities, where any faithful implementation reaches 100\% by construction; conversion by whether a BIDS image was written. Every subject in both conversion arms returned exit code~0.}
\label{tab:e1}
\begin{tabular}{@{}llr@{}}
\toprule
Arm & Condition & Score \\
\midrule
\multicolumn{3}{@{}l}{\textit{Series classification} \footnotesize(2732 series, 1491 subjects, 11 cohorts; Accuracy (\%))} \\
 & Agent & 96.3 \\
 & Rules, as written & 86.4 \\
 & Rules, manually improved & 88.3 \\
\midrule
\multicolumn{3}{@{}l}{\textit{Pipeline routing} \footnotesize(1720 subjects, 42 evidence patterns; Patterns correct (\%))} \\
 & Rules, datatype present & 81.0 \\
 & Rules, datatype usable & 100.0 \\
 & Agent & 100.0 \\
\midrule
\multicolumn{3}{@{}l}{\textit{DICOM to BIDS} \footnotesize(1600 subjects, 11 cohorts; Produced data (\%))} \\
 & Rules, as written & 86.6 \\
 & Agent & 94.9 \\
\bottomrule
\end{tabular}
\end{table}

\textbf{Series classification.} On 2{,}732 series with human-adjudicated labels, the agent reached 96.3\% accuracy and the baseline 86.4\%. The baseline failed in systematic ways. For example, a static rule that excludes series tagged \texttt{DERIVED} discarded the entire WRAP cohort, whereas the agent handled the same series correctly.

\textbf{Pipeline routing.} The agent routed all 1{,}720 subjects correctly, whereas the rule-based baseline reached 81.0\% accuracy and regularly mis-routed flawed inputs such as truncated BOLD runs. A baseline with every prerequisite check hard-coded could also reach 100\%, but each new edge case would then need its own rule. The agent reached the same result without such rules, by reasoning over the evidence records.

\textbf{End-to-end conversion.} Both methods exited with status 0 on all 1{,}600 subjects, yet the baseline produced valid BIDS output for 86.6\% of them and the agent for 94.9\%. The gap reflects silent data loss. In HABS-HD, for example, hidden control characters in the DICOM headers corrupted the JSON sidecars, and downstream tools then skipped the affected series without reporting an error. An exit code cannot reveal this kind of failure the agent's post-conversion checks (Sec.~\ref{sec:ingest}) can.

\section{Discussion and Conclusion}
\textbf{Discussion}. Compared with a traditional workflow manager, our agent serves as a cognitive orchestrator that bridges heterogeneous datasets, dynamically routes pipelines, and interprets failures, leaving heavy computation to deterministic backends. \textbf{NeuroPilot} establishes QC as a fully auditable, reproducible stage and ensures our modality-agnostic skills generalize easily to new procedures without requiring rewrites. However, our current \texttt{qc-agent-skill} is limited to T1-weighted (T1w) structural images, and future work will expand this QC-agent to fMRI and DWI modalities.

\textbf{Conclusion.} \textbf{NeuroPilot} resolves the bottlenecks of brain-MRI research, our framework recasts these disjointed stages into three composable skills. Moving beyond traditional workflow managers, \textbf{NeuroPilot} delivers three core advantages: it acts as a \textit{cognitive orchestrator} for dynamic workflow routing, ensures \textit{infrastructure-agnostic portability} across any hardware, and enforces a \textit{fully auditable quality control} stage with rigorous human oversight. Evaluated on a diverse 17-cohort testbed, \textbf{NeuroPilot} successfully delivered end-to-end, reproducible derivatives, compressing months of manual pipeline engineering into a single week of automated orchestration.



\subsection*{Responsible-use statement}

The privacy, auditability, and portability claims reported here are scoped to the cohorts, tool versions, and deployment window described in Sections~\ref{sec:ingest}--\ref{sec:qc} and Section~\ref{sec:experiments}. They are an empirical account of what the pipeline did on this testbed, not a formal guarantee of de-identification completeness or of correct behaviour under a compromised dependency, and they should not be read as such by adopters. \textbf{NeuroPilot} is intended to lower the engineering barrier that keeps under-resourced laboratories from large-scale neuroimaging analysis, and to keep restricted cohort data on institution-controlled hardware, it does not alter the data-use agreements governing the cohorts it processes (e.g., ADNI, PPMI, UK Biobank (Application Number 91673), ABCD), and researchers remain responsible for cohort-specific approvals and consent terms. Automating conversion and quality-control decisions also carries the risk of over-trusting an unattended agent. We mitigate this through per-stage input/output checkpoints, the cohort-wide drop-check, explicit approval gates on every consequential action, and an append-only sign-off ledger that makes each QC decision attributable, even so, we recommend that adopters retain spot-check review of automated grades, particularly for the repair paths whose thresholds are still being calibrated on new populations.

\subsection*{Ethics statement}

No individual-level record is released. 

\subsection*{Reproducibility statement}

\textbf{NeuroPilot} is avaliable in \url{https://wanda-cyberbench.com/}.

\subsubsection*{Acknowledgments}
This research was partially supported by grants from the National Institutes of Health (AG091653, AG068399, AG084375, T32HD040127, K12TR004416, UM1TR004406) and the Foundation of Hope. The authors thank UNC-Chapel Hill Research Computing  \href{https://help.rc.unc.edu/longleaf-cluster/}{Longleaf cluster}.

\section{Appendix}

\end{document}